\documentclass[runningheads]{llncs}
\usepackage{graphicx}
\usepackage{comment}
\usepackage{amsmath,amssymb} 
\usepackage{breqn}
\usepackage{color}

\usepackage{xcolor}
\usepackage[T1]{fontenc}
\usepackage{adjustbox}
\usepackage{comment}
\usepackage[anythingbreaks]{breakurl}
\usepackage{multirow}
\usepackage{graphbox}

\usepackage{bm}
\usepackage{bbm}
\usepackage{pifont}

\newcommand{\red}{\textcolor{black}}

\newcommand{\RomanNumeralCaps}[1]
    {\MakeUppercase{\romannumeral #1}}

\newcommand\blfootnote[1]{%
  \begingroup
  \renewcommand\thefootnote{}\footnote{#1}%
  \addtocounter{footnote}{-1}%
  \endgroup
}

\begin{document}
\pagestyle{headings}
\mainmatter

\title{One-Shot Object Detection without Fine-Tuning} 

%
\author{Xiang Li\inst{1\thanks{Equal contribution.}} \and
Lin Zhang\inst{1\footnotemark[1]}\and
Yau Pun Chen\inst{1}
\and
Yu-Wing Tai\inst{2}
\and
Chi-Keung Tang\inst{1}}
\authorrunning{F. Author et al.}
%
\institute{HKUST \\
\email{\{xlide, lzhangbj, ypchen\}@connect.ust.hk, cktang@cs.ust.hk}\\
\and
Tencent\\
\email{yuwingtai@tencent.com}}
\maketitle

\begin{abstract}
Deep learning has revolutionized object detection thanks to large-scale datasets, but their object categories are still arguably very limited. In this paper, we attempt to enrich such categories by addressing the one-shot object detection problem, where the number of annotated training examples for learning an unseen class is limited to one. 
We introduce a two-stage model consisting of a first stage Matching-FCOS network and a second stage Structure-Aware Relation Module, the combination of which integrates metric learning with an anchor-free Faster R-CNN-style detection pipeline, eventually eliminating the need to fine-tune on the support images. We also propose novel training strategies that effectively improve detection performance. Extensive quantitative and qualitative evaluations were performed and our method exceeds the state-of-the-art one-shot performance consistently on multiple datasets.

\blfootnote{\red{This research is supported in part by Tencent and the Research Grant
Council of the Hong Kong SAR under grant no. 1620818.}}


\end{abstract}

\section{Introduction}

Modern deep learning models for object detection are trained on
large-scale datasets which are still limited in object categories. Comparing to a rough estimation of 10,000 to 30,000
object categories on the planet Earth~\cite{CMU_grad_vision_slides}, existing large-scale image datasets, such as PASCAL VOC~\cite{voc} (20
classes), ImageNet~\cite{imagenet} (200 classes for detection) and COCO~\cite{coco} (80
classes) are quite limited despite the large number of images in the
respective datasets. For example, in
COCO~\cite{coco} the mean number of images per class is
over 2000 with a high  variance among the 80
classes. Such a large volume of images actually makes it
difficult to extend new object categories
because a large amount of human annotation labor is required. Therefore, enabling the model to learn a novel category given only a few, or even one annotated image becomes especially valuable.

Although transfer learning may be a working solution
for applications with very dissimilar/little data available, 
traditional transfer strategies such as pre-training suffer severely from undesirable biases toward the training class in the final trained models, and fine-tuning the model for each novel class can easily lead to overfitting and reduce the practicality of the method by introducing an extra step each time a new class is learned. These problems are especially severe for the complicated object detection task.

In this paper, we propose a two-stage model for one-shot object detection in the following setting:
given an input with 
a) one support instance patch cropped from the support image by the bounding box of the class of interest 
and b) a query image set without ground truth bounding boxes, without any further change to the model parameters,
the model properly detects object instances in the image and localizes their correct bounding boxes, despite that the relevant object class in the input is 
previously unseen by our trained model. 

In our two-stage model, the first stage Matching-FCOS aims to maximize the recall 
of the bounding box proposals by integrating support information in an anchor-free, metric-learning fashion, while the second stage Structure-Aware Relation Module (SARM) increases the detection precision with classification and box regression. Divided into two stages, our model also benefits from our proposed training strategies that enhance different parts of the training process, namely, 1) Query-Support Feature Similarity Mining, 2) Groundtruth-Curated Proposals, and 3) Second Stage Knowledge Transfer. 
With all the above combined, our method can be easily extended to unseen classes given one support instance patch as input and without any fine-tuning.
Our experimental results show that our model achieves state-of-the-art 
results on PASCAL VOC dataset~\cite{voc} under the one-shot setting without fine-tuning, performing significantly better than LSTD~\cite{lstd} and Repmet~\cite{repmet}, the representative state-of-the-arts where fine-tuning is applied, while also outperforming one of the most recent method CoAE~\cite{co-ae}.

\vspace{-0.2in}

\section{Related Work}

\vspace{-0.1in}

\noindent {\bf Few-shot Learning and Classification.~~} 
In few-shot learning setting, a model is pre-trained on the data with abundant labeled classes, and aims to predict the result for unseen, novel classes given only a few labeled examples. Few-shot learning research has mainly been focused on image classification. 

In the recent research, there are mainly three categories of few-shot classification approaches: 1) train a good initial network to be fine-tuned on extremely small training set, as proposed in~\cite{modelag,op}; 2) exploit recurrent neural network's (RNN) memory properties, as discussed in~\cite{metanetwork,memoryaug}; 3) learn a metric between few-shot samples and queries, as in~\cite{feedforward,prototype,omniglot,siamese,ltc,matchnet}.

One example of the first category is~\cite{modelag} which meta-learns a good initialization of the network, enabling it to be fine-tuned on novel classes with only a few support examples. 
The other approach learns an RNN-based optimizer for the network, so that novel classes can be classified referencing previous information from the RNN memory accumulated during training. This line of work is represented by~\cite{metanetwork} and~\cite{memoryaug}. 
Finally, one of the most successful approaches to few-shot learning in recent years is metric learning. This family of methods attempts to learn a projection from the image space to a feature space, where the image features are classified with different methods ranging from a simple nearest-neighbor~\cite{prototype} to a learned neural network classifier~\cite{ltc}. 

Our method primarily takes inspiration from metric learning and~\cite{matchnet}, where the support information is also integrated in a simple feed-forward way. 
Also, our network does not require fine-tuning which is necessary to all existing meta-learning methods.


\noindent {\bf Object Detection.~~}
Extensive work has been done on deep learning-based object detection. 
\red{
Two-stage object detection methods, represented by Faster R-CNN~\cite{fasterrcnn}, are commonly comprised of a class-agnostic region proposal network (RPN) to propose regions of the given image that are likely to contain objects, and a multi-headed network that classifies and refines these region proposals to produce the final detection results. 
}
Compared to one-stage methods, two-stage methods usually enjoy better performance because the RPN filters out a large number of irrelevant image features; however, it can also potentially filter out crucial regions when applied in the few-shot learning setting. Previously, detectors employed the anchor mechanism, which uses a large number of predefined potential boxes as a prior, as proposed in~\cite{fastrcnn}. However, this brings a large number of hyperparameters related to the anchors that are data-dependent, which is especially undesirable under the few-shot setting. 

Recently proposed anchor-free methods, e.g. FCOS~\cite{tian2019fcos}, aim to rid the detector of these hyper-parameters
through multi-scale,  per-pixel  regression along  with  an  auxiliary artificial training target named “centerness”, while also achieving comparable performance to anchor-based methods. 
Our model consists of a two-stage network architecture, with its first stage based on FCOS~\cite{tian2019fcos}, and the second stage similar to Faster R-CNN \cite{fasterrcnn}.

\noindent {\bf Few-shot Object Detection.~~} Only a limited number of previous work has focused on the few-shot object detection task. In LSTD~\cite{lstd} the authors proposed a few-shot detector by transfer learning. With regularization mechanisms to reduce the domain difference between training classes and test classes, this work aims to detect novel class objects by pre-training on the large-scale training set and then fine-tuning on the novel class test set. RepMet~\cite{repmet} differs from the previous approach by applying a metric learning classification model directly after a traditional region proposal network. Our method develops upon the two previous methods in that our model does not require additional fine-tuning for each novel class, and we propose to not only integrate the support information to the second stage, but also to the region proposal of the first stage, which can improve the first stage recall by a large margin as shown in our experiments.

Recently, many new few-shot object detection methods have emerged: in~\cite{feature-reweighting} base class predictions are re-weighted to generate novel class predictions in one stage. ROI meta-learning was applied in~\cite{meta-rcnn}, while in~\cite{co-ae} co-attention and co-excitation were used, and attention-RPN, multi-Relation detector and contrastive training were applied in~\cite{qifan} which used more than one shot.



\begin{figure*}[t]
\begin{center}
   \includegraphics[width=1.0\linewidth]{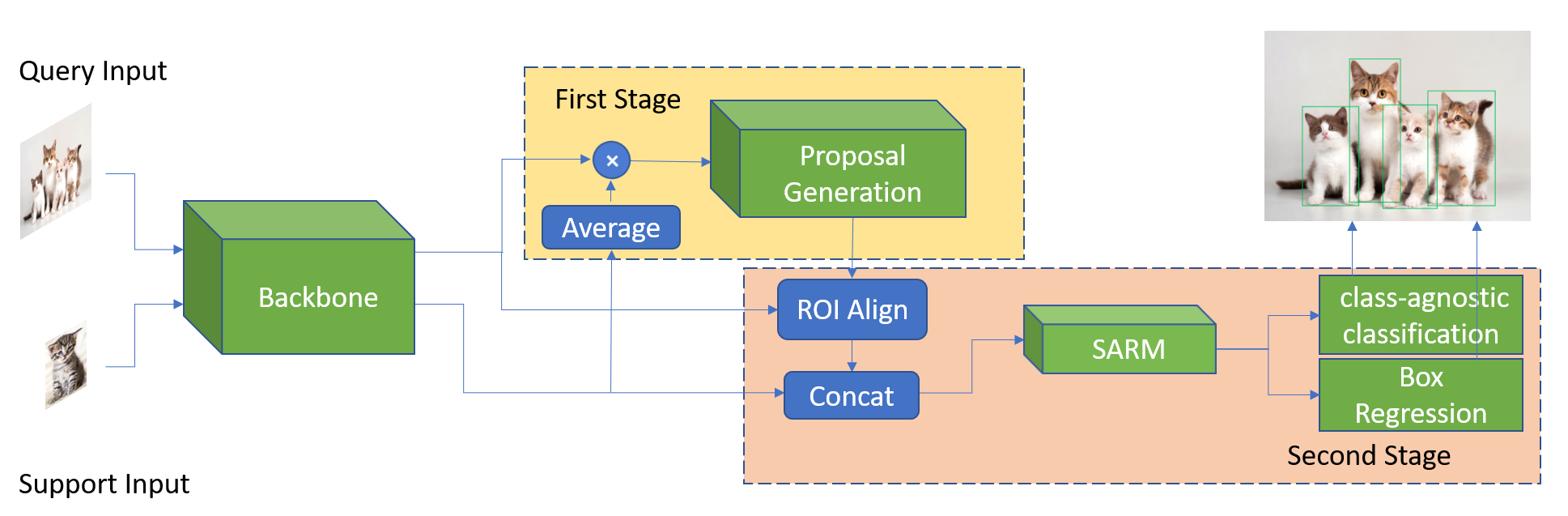}
\end{center}
\vspace{-0.2in}
   \caption{Overview of our architecture. The query image and support instance patch are first fed into a shared siamese backbone network. Then our Matching-FCOS produces a set of high-recall proposals. The second stage, which we term Structure-Aware Relation Module (SARM), learns to classify and regress bounding boxes by focusing on structure local features. The final goal is to detect objects in the query image with the same class of the support object. }
\label{overview}
\vspace{-0.18in}
\end{figure*}

\vspace{-0.15in}

\section{Method}

In this work, we design both the neural network architecture, as shown in Figure~\ref{overview}, and our training and inference methods. We build a one-shot object detection network by modifying the FCOS~\cite{tian2019fcos}, a fully convolutional, one-stage, anchor-free object detection network. Using the FCOS backbone makes the network anchor-free, unlike most region proposal network RPN-based work~\cite{fasterrcnn}, as anchor mechanism such as Faster R-CNN~\cite{fasterrcnn} and SSD~\cite{ssd} are seen as less effective~\cite{Zhuoyao}. We combine FCOS~\cite{tian2019fcos} with one-shot learning by introducing a Siamese backbone, taking inspiration from~\cite{siamese},  so that the support and query images were both input to the same Feature Pyramid Network~\cite{fpn}  to produce their corresponding feature representations. 
Then, the support features undergo a global average pooling layer to produce a global representation of the support instance. This pooled feature is then convolved with the query feature map, which essentially computes the dot product between the support and query representations to obtain a similarity map.
Finally, this similarity map is fed to the following head networks that are identical to the original FCOS~\cite{tian2019fcos}. The loss is calculated between the prediction of the network and the ground-truth boxes of the support class in the query image annotation, where back-propagation is applied to optimize the network parameters. 

In the following, we describe our method in detail. 

\subsection{One-shot Object Detection}
Following the episode-based formulation in~\cite{ltc}, the dataset is split into into two disjoint subsets by object category, namely the training classes $\bold{C}_{train}$ with data $\bold{D}_{train}$ and test classes $\bold{C}_{test}$ with data $\bold{D}_{test}$. Each of them is further divided into the disjoint query set $\bold{Q}$ and support set $\bold{S}$, i.e.,

\begin{align*}
\bold{C}_{train} \cap \bold{C}_{test} &= \varnothing \\
\bold{D}_{train} = \bold{Q}_{train} \cup \bold{S}_{train} &; 
\bold{Q}_{train} \cap \bold{S}_{train} = \varnothing \\
\bold{D}_{test} = \bold{Q}_{test} \cup \bold{S}_{test} &; 
\bold{Q}_{test} \cap \bold{S}_{test} = \varnothing \\
\end{align*}





Specifically, in our one-shot detection problem, $\bold{S}_{test}$ contains a labeled support instance patch for each test class (a \textit{shot}) and $\bold{Q}_{test}$ contains a number of unlabeled query images for each test class.


In each training iteration, the input to the model is a collection of support and query images, e.g., the input 
$ \bold{I} = \{ s, q \}$,
where $s \in \bold{S}_{train}$ represents a support instance patch cropped from a given support image containing a specific class $c \in \bold{C}_{train}$, and $q \in \bold{Q}_{train}$ represents a query image.
The target of the model is to predict $l_q^c$, the bounding boxes for class $c$ objects in the query image $q$. The setting is similar for test iterations. 

Following our problem settings, in each training and testing step, the model assumes no prior information of either the support or query class. Through the training process, the network learns to compare the potential regions of the query image and the support instance patch instead of accumulating class-specific knowledge. Therefore, in practice, in order to obtain
more different training support-query pairs, we do not apply the restriction of $\bold{Q}_{train}$ and $\bold{S}_{train}$ to be disjoint. Instead, we only need to ensure in each iteration, $s$ and $q$ are distinct images. However, we always make sure that the training classes and the test classes are disjoint in order to evaluate the model's ability to generalize to unseen novel classes.

Note that the above setting can be readily extended to address the N-way K-shot detection problem, where N novel classes are detected at the same time and K support instance patches are available for each class. Although this is not our paper's focus, we present the straightforward extension to N-way K-shot setting in the supplementary material.





\subsection{Model}

We propose a novel, effective and class-agnostic network architecture for one-shot object detection in two stages. The first stage, named matching-FCOS, aims to generate object proposals with a high recall by integrating the information from the support images; while the second stage, with the Structure-Aware Relation Module, further classifies the object proposals and applies a box regression to produce the final detection result.  Figure~\ref{overview} shows the two-stage pipeline.

\begin{figure}[t]
  \centering
  \includegraphics[width=.8\linewidth]{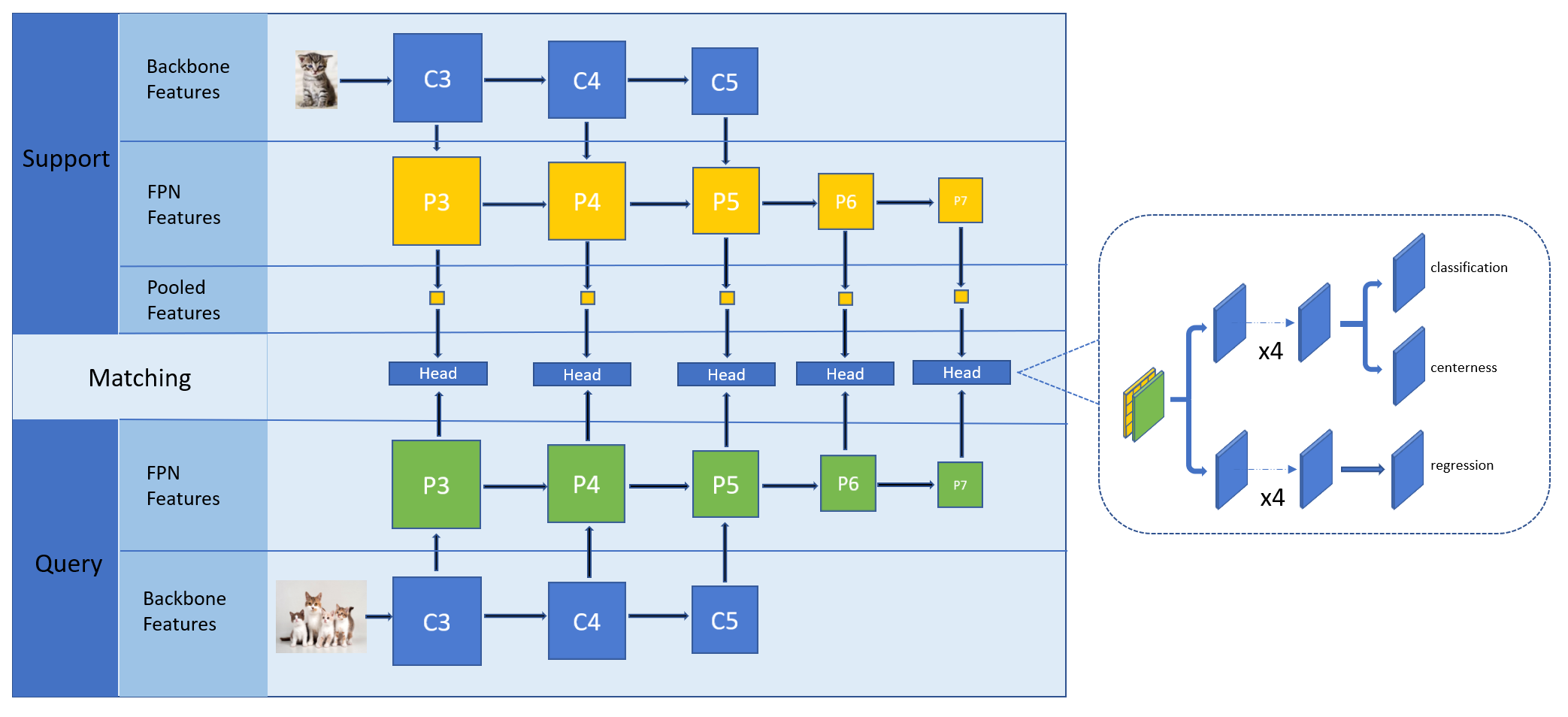}
  \vspace{-0.1in}
  \caption{  Matching-FCOS network as the first stage of our model. C3--C5 refers to feature maps of the backbone and P3--P7 refers to feature maps of the FPN. }
  \label{fig:matching-fcos}
  \vspace{-0.2in}
\end{figure}

\begin{table}[t]
\begin{center}
\begin{adjustbox}{max width=0.99\linewidth}
\begin{tabular}{l|c|c|c}
\hline
Model  & AR\textsubscript{100} & AR\textsubscript{1000} & AR\textsubscript{4000} \\
\hline
RPN             & 65.4 & 88.4 & 91.5 \\
FCOS            & 69.5 & 91.3 & 92.7\\
Matching-FCOS   & \textbf{80.7} & \textbf{94.6} & \textbf{95.3} \\
\hline
\end{tabular}
\end{adjustbox}
\end{center}
\vspace{-0.05in}
\caption{  Comparison of the recall for different first stage methods. The IoU threshold is set to 0.5.}
\vspace{-0.3in}
\label{compRecall}
\end{table}

\vspace{0.05in}

\noindent {\bf Matching-FCOS.~~}
Figure~\ref{fig:matching-fcos} shows the structure of our Matching-FCOS.
Traditional RPN~\cite{fastrcnn} struggles in one-shot object detection since the generated proposals tend to bias toward the training class objects, while objects of the novel class are often overlooked especially when their appearances are vastly different from the training class objects. 

Such objects overlooked in the first stage are often next to impossible to retrieve in the second stage, as the second stage bounding box regression only models comparatively small discrepancies due to its linear approximation nature. 
Table~\ref{compRecall} shows the recall of traditional RPN setting a low performance upper bound to the detection process. A more effective structure, i.e., FCOS~\cite{tian2019fcos} is beneficial but the improvement is limited. We believe that it is crucial to introduce metric learning to the support generation process to enable the network to learn to generate novel class proposals by comparison in order to improve the recall in the first stage. Therefore, we propose the Matching-FCOS which fuses the multi-scale support and query information for few-shot proposal generation.

First, the support instance patch and the query image are passed through a weight-shared feature pyramid network~\cite{fpn} in a siamese fashion to produce the support and query feature. Then we obtain a global representation of the support instance patch in each scale by average pooling over the support feature map. The average-pooled support features and the query features from the FPN are passed into the heads for matching at each scale.
Each head computes the cosine similarity between each pixel of the query features and the global representation of the support features to generate a similarity map. Finally, following the original FCOS head structure, pixel-wise classification and box regression are applied to the similarity map to produce the box proposals.

Note that Matching-FCOS generates proposals in an anchor-free fashion, which eliminates the need to tune hyper-parameters concerning anchors in the one-shot evaluation, thus enabling the network to generalize better in the proposal generation of novel classes given sufficient variety in the training dataset. The effectiveness of incorporating support information to the proposal generation of the Matching-FCOS is validated by our experiments in Table~\ref{compRecall}, which demonstrate a significantly higher recall than traditional RPN~\cite{fasterrcnn} and the traditional FCOS, especially for AR\textsubscript{100}.
Additionally, despite that the Matching-FCOS is designed to obtain high recall, it can also be readily used as a one-stage one-shot detector with a moderate performance loss. 
Refer to the experimental section for details.

\begin{figure}[t]
  \centering
  \includegraphics[width=.95\linewidth]{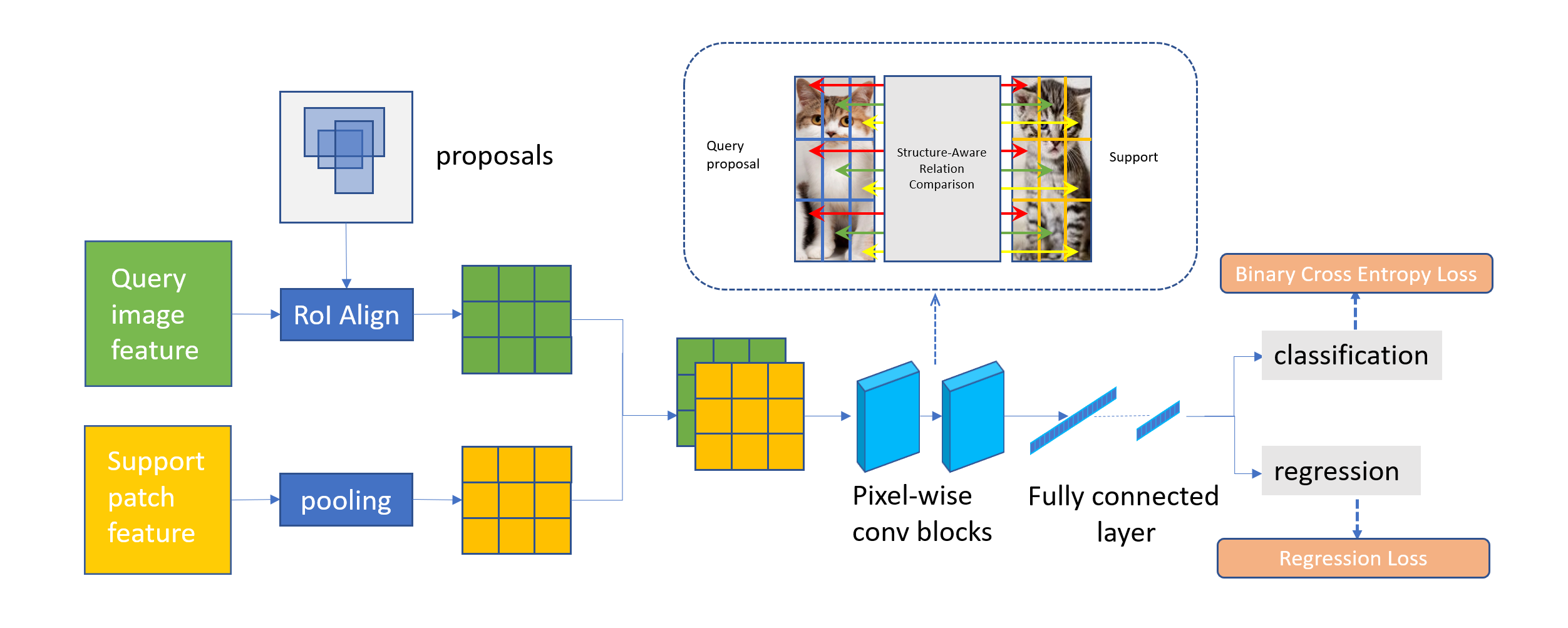}
  \vspace{-0.15in}
  \caption{  Structure-Aware Relation Module at the second stage. We first pool query proposals features and support features into $K \times K$ features and concatenate them. We then use pixel-wise convolutional layers to compare structure-aware local features. Here, the cat is decomposed into structural modules including ears, feet, etc. By processing these features locally, our module can discover more relevant cues to achieve higher detection precision. }
  \label{fig:imagecompare}
\end{figure}


\begin{figure}[t]
\vspace{-0.15in}
\centering
\begin{minipage}{\textwidth}
\begin{minipage}[b]{.42\textwidth}
    \begin{center}
    \begin{tabular}{l|c|c}
    \hline
    Model  & AP\textsubscript{50} &  AP\textsubscript{75} \\
    \hline
    Global-average Head         & 41.8   & 23.0 \\
    Faster R-CNN Head            & 45.3   & 22.1 \\
    SARM                        & \textbf{46.1} & \textbf{23.7}  \\
    \hline
    \end{tabular}
    \end{center}
    \vspace{-0.05in}
    \caption{  Comparison of different head structures at the second stage in Task~\RomanNumeralCaps{2}.}
    \vspace{-0.15in}
    \label{compHead}
\end{minipage}%
\hfill
\begin{minipage}[b]{.55\textwidth}
  \centering
  \includegraphics[width=.95\linewidth]{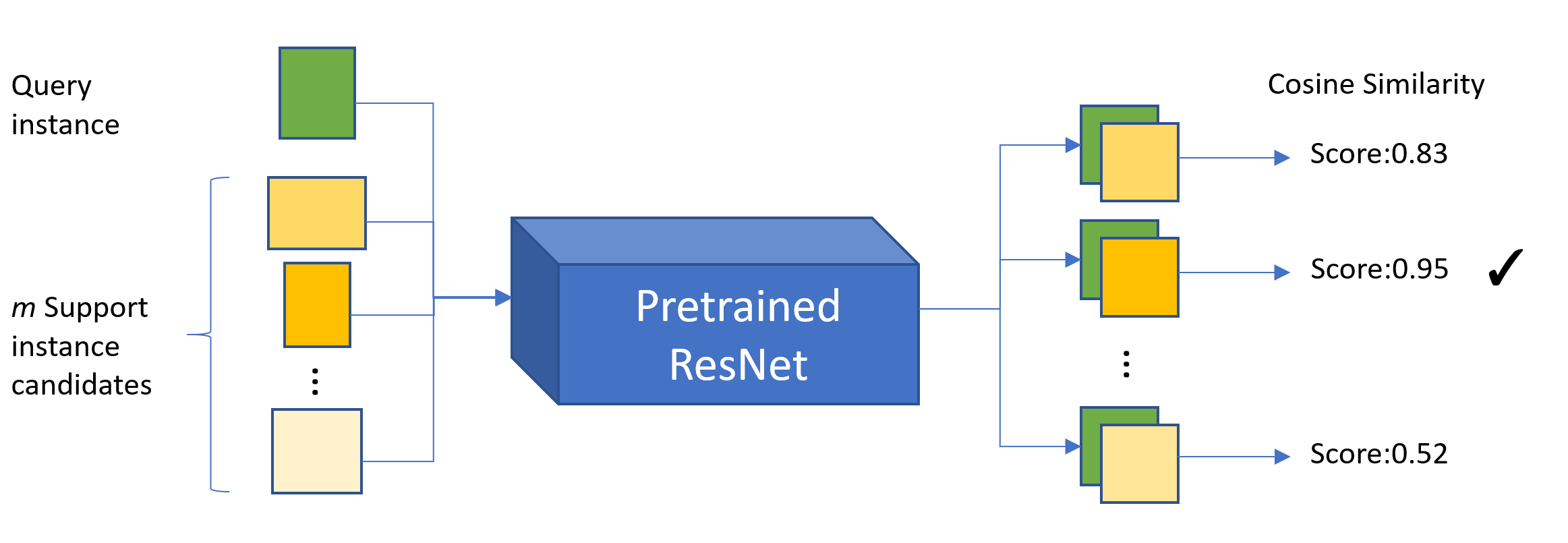}
  \vspace{-0.05in}
  \caption{ \red{Illustration of Query-Support Feature Similarity Mining. }}
  \label{fig:sim-mining}
  \vspace{-0.15in}
\end{minipage}
\end{minipage}
\end{figure}


\vspace{0.05in}

\noindent {\bf Structure-Aware Relation Module.~~}
Figure~\ref{fig:imagecompare} illustrates the structure of the second stage SARM model.
Since the Matching-FCOS targets at obtaining a high recall, the support features were globally averaged pooled, where the spatial and structural information in the support instance patches are lost. As shown in the first row of Table~\ref{compHead}, a second stage model that directly uses the globally averaged pooled support features from the first stage proves detrimental to the detection performance, demonstrating a large proportion of false-positive detections and poor localization. 

To preserve such local, structural information in the support features, the second row in Table~\ref{compHead} shows a standard Faster R-CNN head modified with relation network~\cite{sung2018learning} to incorporate support features without explicitly enforcing a spatial comparison of support and query features. Although the AP\textsubscript{50} achieves a large improvement, AP\textsubscript{75} has dropped, suggesting worse detailed localization.
Therefore, we believe it is not only necessary to compensate for the loss of support structural information in the first stage, but also to draw a local, structure-aware feature space comparison of such support features and the features of query proposal region, so that such information can be infused more effectively at the second stage.

Our Structure-Aware Relation Module (SARM) is designed to accomplish both goals to fully utilize the structural and local information in the support instance patches, and to learn the structural correlation between support and query images. This is achieved by decomposing the image feature maps to compare the corresponding local features in our SARM module.  Table~\ref{compHead} shows that this outperforms both the previous second stage structures.

We first crop support and query features into $N{\times}N$ fixed-size feature maps with ROI Align operation separately~\cite{maskrcnn}. Accordingly, both the support and query feature map are evenly divided where each partition contains rich information of its surrounding local features. This structured feature map thus retains the support object's local features to a larger extent. Then, a series of pixel-wise convolution blocks are added to enforce the network to learn to compare local features of the support and query feature maps. Such a structure not only enforces the network to learn the local correlation of the support and query features, but also restricts the comparison of unrelated parts in the features space. Finally, the combined feature map is flattened and passed through fully connected layers for classification and regression, which collects the local information to make an informed global decision while preserving local context.


The pixel-wise convolution blocks that enable SARM to achieve better performance shares similarity with the attention mechanism, which explicitly models the relation between feature pixels commonly using convolutional layers~\cite{yu2018generative}. However, instead of capturing long-range dependencies as in most previous work on attention, we aim to restrict the network from comparing distant features by focusing on each local region.

\noindent {\bf No extra fine-tuning.~~}
Different from previous works represented by LSTD~\cite{lstd}, upon encountering a novel class, our model can be applied directly without any modification, taking the query images and support patches as input and outputting detection results without  fine-tuning the network on the given support images for the new class. Our experiments show that without fine-tuning, our method can even achieve higher performance than the state-of-the-art in the one-shot setting, rendering it much more preferable in real-world applications.

\vspace{-0.2in}

\subsection{Loss Functions}

\noindent {\bf First Stage Loss.~~} In the first stage, for each bounding box, we predict a 2D classification vector $\bold{p}\in[0, 1]$ representing the probability for background and target object respectively and a location vector $\bold{t} = (l, t, r, b)$ where $l,t,r,b$ refer to the distances from the location to the left, top, right and bottom of the ground truth box. 
Following \cite{tian2019fcos}, we tackle the imbalance of target and background objects by applying focal loss for classification, and we choose IOU loss in UnitBox \cite{Unitbox} for box regression. In combination:
\begin{align*}
    {\cal L}_{first} = \frac{1}{N_{pos}} \sum_{x,y}{\cal L}_{focal}(\bold{p}_{x,y},  c^{*}_{x,y}) 
    + \frac{1}{N_{pos}} \sum_{x,y}\mathbbm{1}_{\{c^{*}_{x,y}=1\}} {\cal L}_{unitbox}(\bold{t}_{x,y}, \bold{t}^{*}_{x,y}) 
\end{align*}
where $x,y$ refer to locations, $N_{pos}$ is the number of positive examples, $c^*$ refers to class labels and $t^*$ refers to the groundtruth regression target.

\noindent {\bf Second Stage Loss.~~} The second stage output includes a classification vector $\bold{p} \in [0, 1]$ for background and target object and a regression vector $\bold{t}$. Different from stage one, $\bold{t}=(t_{x}, t_{y}, t_{w}, t_{h})$, following the parameterization in~\cite{rcnn}, specifies a scale-invariant translation and log-space height/width shift relative to an object proposal. In the second stage, we adopt binary cross entropy loss for classification and smooth L1 loss for regression following \cite{fastrcnn}:
\begin{align*}
    {\cal L}_{second} = \frac{\lambda_{cls}}{N} \sum_{i}\mbox{BCE}(\bold{p}_{i},  c^{*}_{i}) 
    + \frac{1}{N_{pos}} \sum_{i}\mathbbm{1}_{\{c^{*}_{i}==1\}} {\cal L}_{smoothL1}(\bold{t}_{i} -  \bold{u}^{*}_{i}) 
\end{align*}
where $\lambda_{cls}=2$ is a balancing factor.

Our final loss function is the sum of the first and second stage loss:
\begin{equation*}
    {\cal L} = {\cal L}_{first}+\lambda_{second} {\cal L}_{second}
\end{equation*}
where $\lambda_{second}$ is the balancing factor for second stage. In our experiments we choose $\lambda_{second}=2.5$.

\subsection{Query-Support Feature Similarity Mining}



In our metric learning-based method, the detection relies on the similarity of the support and query object in a deep embedding space, thus the support instance patches play a key role in guiding the detection. However, in current large-scale datasets, e.g., COCO~\cite{coco}, even for objects in the same category, there still exists a major disparity between different instances. This intra-class variance can be due to the objects' color and pose as well as occlusion, lighting, and background, which introduce unpredictable noises in the training of the relational comparison in both the first stage and the second stage. Therefore, we adopt the strategy of query-support feature similarity mining in the training process.

\red{The Query-Support Feature Similarity Mining process is illustrated in Figure~\ref{fig:sim-mining}. }Specifically, we first crop out all instances in the training dataset based on the ground-truth \red{bounding box} label. In each iteration, one image patch is regarded as a query patch and $m$ other \red{different patches of the same class} are randomly selected as potential support instances, where $m$ is a hyperparameter set to 100 in our experiments. 
\red{We extract the feature map of the one query patch and $m$ support patches with a ResNet-50~\cite{resnet} pretrained with classification task on the entire training set. Next, the cosine similarity is calculated between each pair of support and query features to produce a similarity score:}
\begin{equation*}
    \cos(f_{s}, f_{q}) = \frac{f_{s} \cdot f_{q}}{|f_{s}||f_{q}|}
\end{equation*} 
\red{Finally, for each query image that may contain multiple instances, we ensure the $m$ potential support patches for each of these instances are chosen to be the same. 
Therefore, the average similarity score of the same $m$ potential support objects is considered as the similarity between the query images and the $m$ potential supports, with which we choose the most similar support instance as this query image's support instance patch in training. Note that this process can be done offline to save the training time.}

Following this method, mining similar support patches for training can alleviate the intra-class variance problem and provides a more reasonable learning target, i.e., distinguishing query objects similar to the support objects, for both the first-stage and the second-stage network. \red{The effectiveness of Query-Support Feature Similarity Mining is discussed in the ablation study in Section~\ref{ablation}.}
\begin{table}[t] 
\begin{center}
\begin{adjustbox}{max width=0.99\linewidth}
\begin{tabular}{l|c|c}
\hline
Model &  Task I    &
Task II \\
\hline
 LSTD    & 19.2  &  34.0  \\
                          Repmet &  24.1  &  -  \\
                 Ours  &   \textbf{31.0\textsuperscript{\textbf{+6.9}}} &   \textbf{46.1\textsuperscript{\textbf{+12.1}}} \\ \hline
\end{tabular}
\end{adjustbox}
\end{center}
\vspace{-0.05in}
\caption{  Comparison of \red{mAP\textsubscript{50}} with LSTD~\cite{lstd} and Repmet~\cite{repmet} under 1-shot setting on task~\RomanNumeralCaps{1} and~\RomanNumeralCaps{2}.}
\vspace{-0.15in}
\label{compWithLSTD}
\end{table}


\begin{table}[t] 
\begin{center}
\begin{adjustbox}{max width=0.99\linewidth}
\begin{tabular}{c|c|c|c|c|c|c|c|c|c|c|c|c|c|c|c|c|c|c|c|c|c|c}
\hline
\multirow{2}{*}{Model}  & \multicolumn{17}{|c|}{seen class} & \multicolumn{5}{|c}{unseen class} \\
&plant& sofa& tv& car& bottle& boat& chair& person& bus& train& horse& bike& dog& bird& mbike& table& mAP& cow& sheep& cat& aero& mAP \\
\hline
SiamFC & 3.2 & 22.8 & 5.0 & 16.7 & 0.5 & 8.1 & 1.2 & 4.2 & 22.2 & 22.6 & 35.4 & 14.2 & 25.8 & 11.7 & 19.7 & 27.8 & 15.1 & 6.8 & 2.28 & 31.6 & 12.4 & 13.3  \\
SiamRPN & 1.9 & 15.7 & 4.5 & 12.8 & 1.0 & 1.1 & 6.1 & 8.7 & 7.9 & 6.9 & 17.4 & 17.8 & 20.5 & 7.2 & 18.5 & 5.1 & 9.6 & 15.9 & 15.7 & 21.7 & 3.5 & 14.2  \\
CompNet & 28.4 & 41.5 & 65.0 & 66.4 & 37.1 & 49.8 & 16.2 & 31.7 & 69.7 & 73.1 & 75.6 & 71.6 & 61.4 & 52.3 & 63.4 & 39.8 & 52.7 & 75.3 & 60.0 & 47.9 & 25.3 & 52.1 \\
CoAE & 30.0& 54.9& 64.1& 66.7& 40.1& \textbf{54.1}& 14.7& \textbf{60.9}& 77.5& 78.3& \textbf{77.9} & 73.2& 80.5& 70.8& \textbf{72.4}& 46.2& 60.1& \textbf{83.9}& 67.1& 75.6& 46.2& 68.2 \\
Ours  &\textbf{33.7}&\textbf{58.2}&\textbf{67.5}&\textbf{72.7}&\textbf{40.8}&48.2&\textbf{20.1}&55.4&\textbf{78.2}&\textbf{79.0}&76.2&\textbf{74.6}&\textbf{81.3}&\textbf{71.6} &72.0&\textbf{48.8}& \textbf{61.1} &74.3& \textbf{68.5} & \textbf{81.0} & \textbf{52.4} & \textbf{69.1} \\ 
\hline
\end{tabular}
\end{adjustbox}
\end{center}
\vspace{-0.05in}
\caption{  Comparison of \red{AP\textsubscript{50}} with CoAE~\cite{co-ae} on Task~\RomanNumeralCaps{3}.}
\vspace{-0.3in}
\label{compWithCOAE}
\end{table}

\begin{table*}[t]
\begin{center}
\begin{adjustbox}{max width=0.99\linewidth}
\begin{tabular}{l||c|c|c|c|c|c|c|c|c|c}
\hline
Class & plane & bike & bird & boat & bottle & bus & car & cat & chair & cow \\
\hline
AP\textsubscript{50} & 28.4 & 26.8 & 50.8 & 19.6 & 18.4 & 67.5 & 64.0 & 69.5 & 26.3 & 84.8 \\
\hline
 Class & table & dog & horse & mbike & person & plant & sheep & sofa & train & tv \\
\hline
AP\textsubscript{50} &  11.9 & 73.7 & 67.7 & 26.4 & 9.6 & 10.3 & 75.1 & 79.1 & 55.4 & 54.2\\
\hline
\end{tabular}
\end{adjustbox}
\end{center}
\vspace{-0.05in}
\caption{  Per-class performance of our method on task~\RomanNumeralCaps{2} under the 1-shot setting, showing that for classes more similar to the training set our method receives better performance while performing less well for very dissimilar classes such as person.}
\vspace{-0.25in}
\label{perClass}
\end{table*}

\subsection{Groundtruth-Curated Proposals}


Through experiments, we have found that despite the high recall of the proposals generated by Matching-FCOS, the proposals often have low IoU with the ground-truth bounding box. 

Low-quality positive proposals generated by the first stage can produce an adverse effect on the second stage classification during training. With these more ambiguous and less well-defined proposals, characteristic features of the object for its correct classification may not be included in the proposal box at all, while a large proportion of the background may be included. This problem has an especially severe impact because the second stage classification is formulated as a similarity comparison between the support and the query in the embedding space under the one-shot setting. 

In order to produce a more conducive training condition for the second stage one-shot classification, we propose to sample boxes with certain IoU thresholds with the ground-truth box and add these proposals curated by the ground-truth to the real proposals generated by the Matching-FCOS as the input for the second stage in training, with the testing unchanged. Our ablation study in the experimental section shows that this strategy is useful.

\vspace{-0.15in}
\subsection{Second Stage Knowledge Transfer} \label{fss-train}


Due to the limited number of object categories in the current object detection datasets, the training data, especially for the second stage classifier network lacks variety, leading to worse classification performance on novel classes that are vastly different from training classes. 

On the other hand, many new image classification or segmentation datasets, e.g.,~\cite{FSS1000} have a large number of categories and variety. But they may not be appropriate for directly training the detection model  since these datasets may lack bounding box annotation. Even if the bounding box annotation is present, the images in these datasets are often dominated by large foreground objects with a clear distinction from the background and the number of instances is limited~\cite{FSS1000}, both of which have an adverse effect on the training of the detection model. 
Therefore, we propose to transfer the knowledge of ample class variety
from classification/segmentation datasets with a large number of categories by pretraining the
second stage Structure-Aware Relation Module alone for the classification task using \red{the instance patches cropped from} these datasets, and then fine-tune the entire model for object detection using existing object detection datasets~\cite{voc,coco}.

\section{Experiments}

\subsection{Comparison with State-of-the-art}

\begin{figure}[t]
\begin{center}
  \includegraphics[align=c,width=0.15\linewidth]{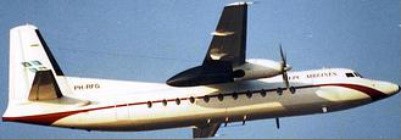}
  \hfill
  \includegraphics[align=c,width=0.15\linewidth]{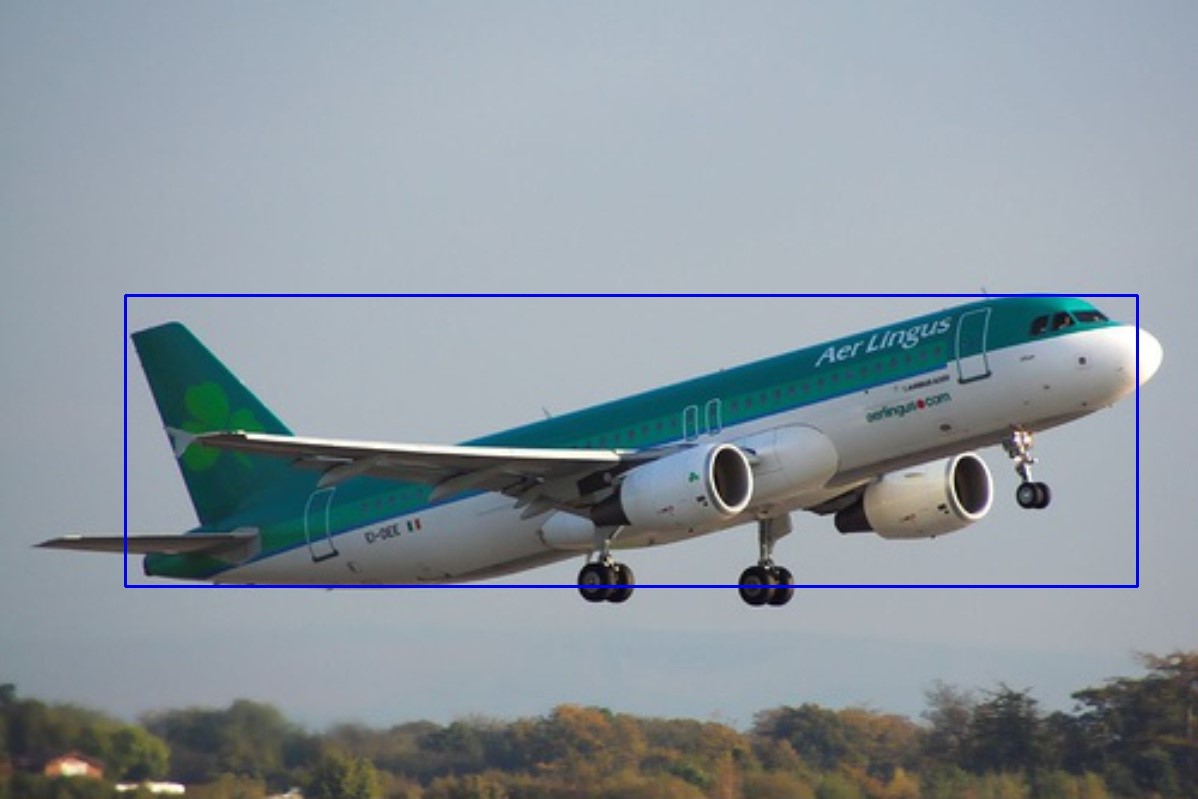}
  \includegraphics[align=c,width=0.15\linewidth]{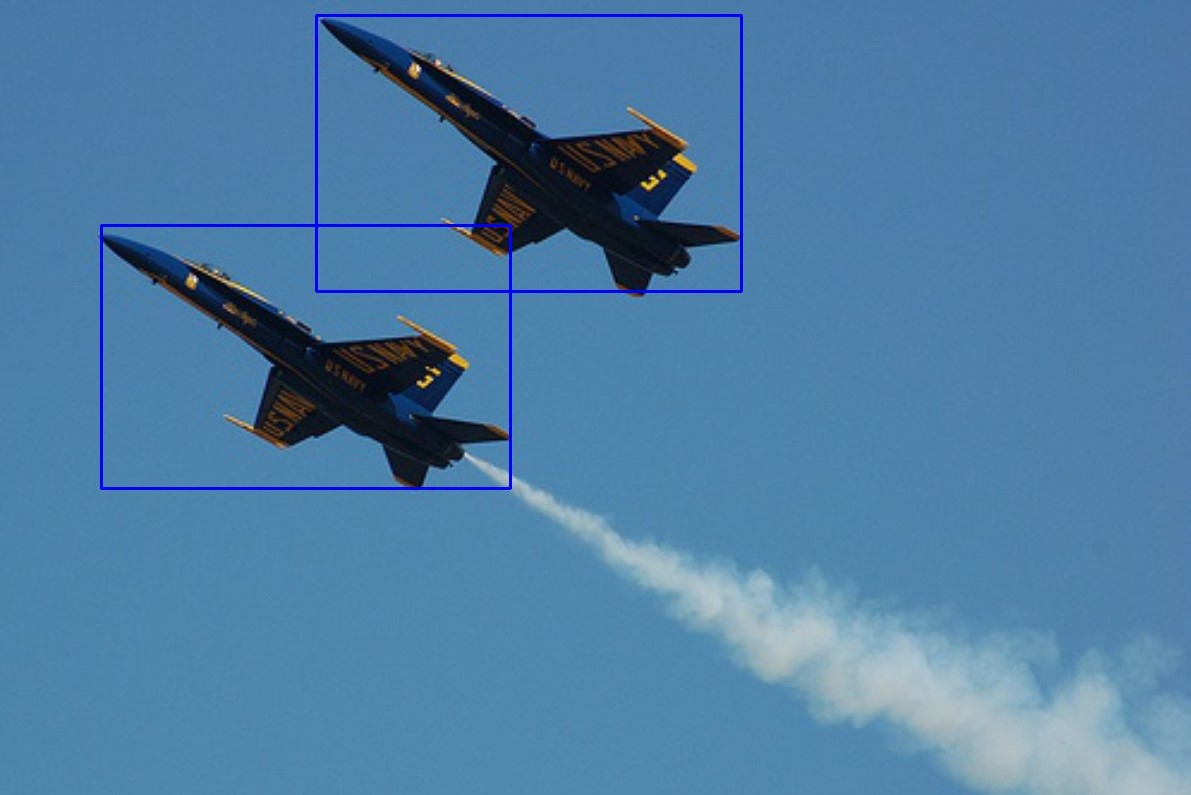}
  \includegraphics[align=c,width=0.15\linewidth]{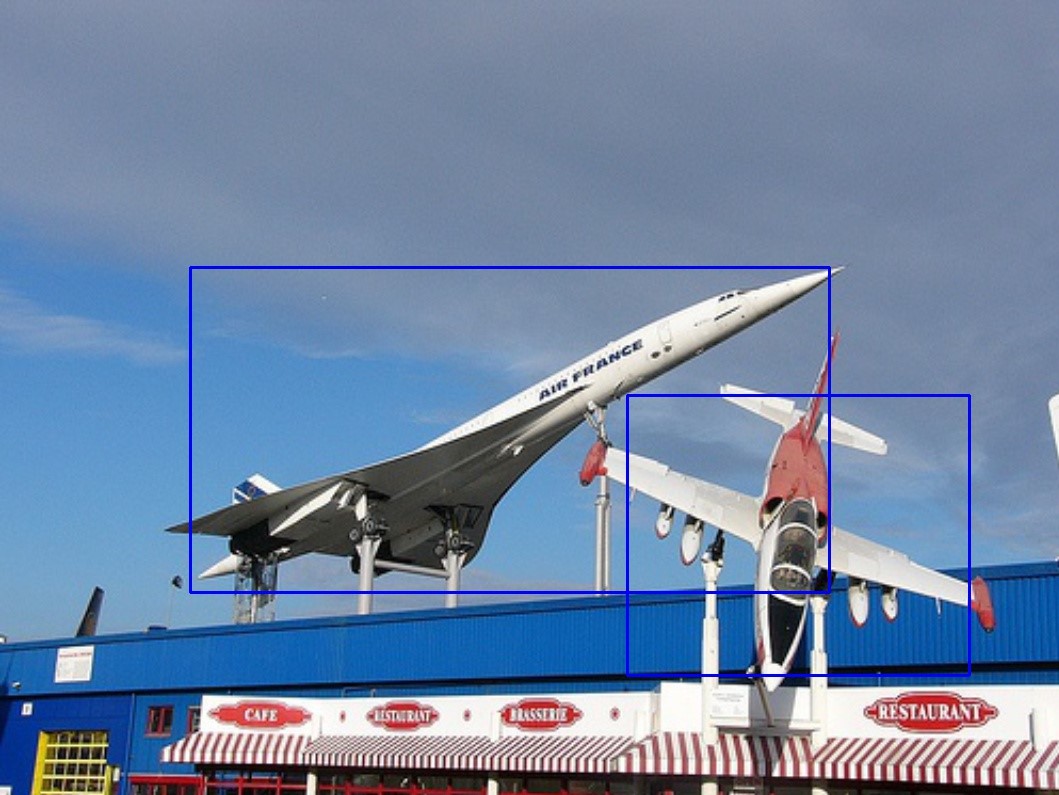}
  \includegraphics[align=c,width=0.15\linewidth]{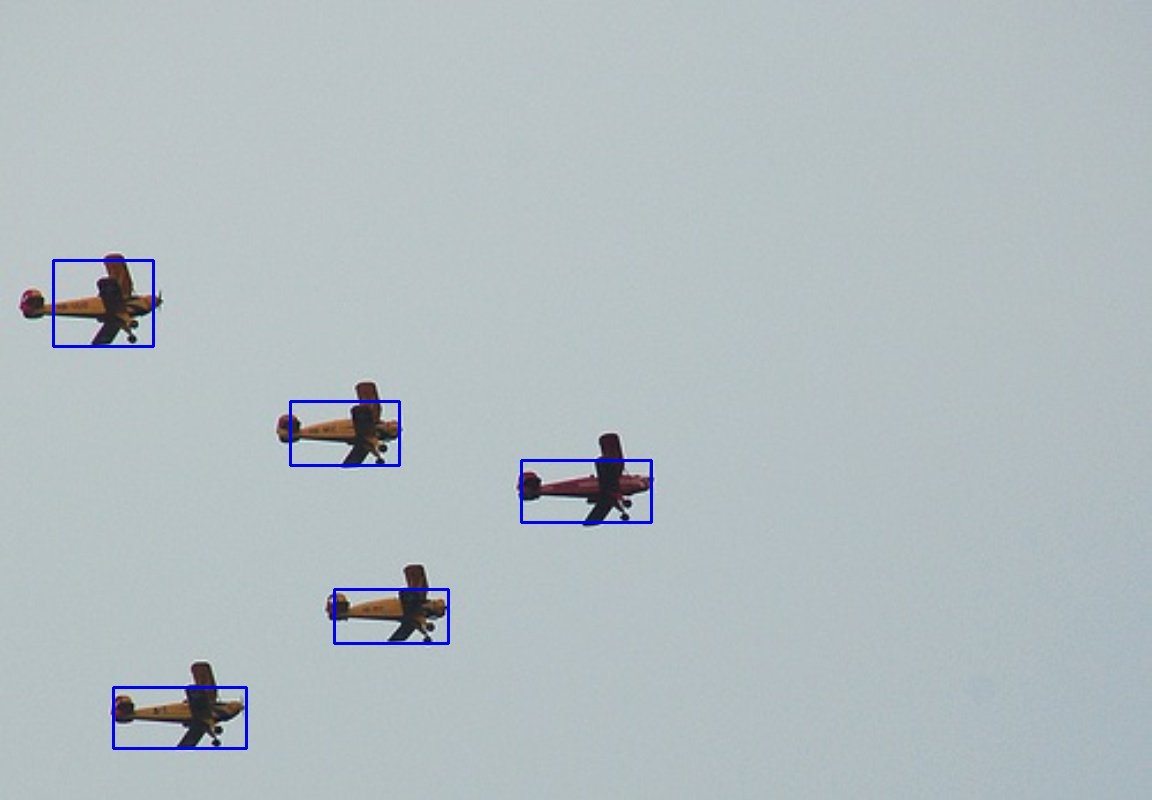}
  \includegraphics[align=c,width=0.15\linewidth]{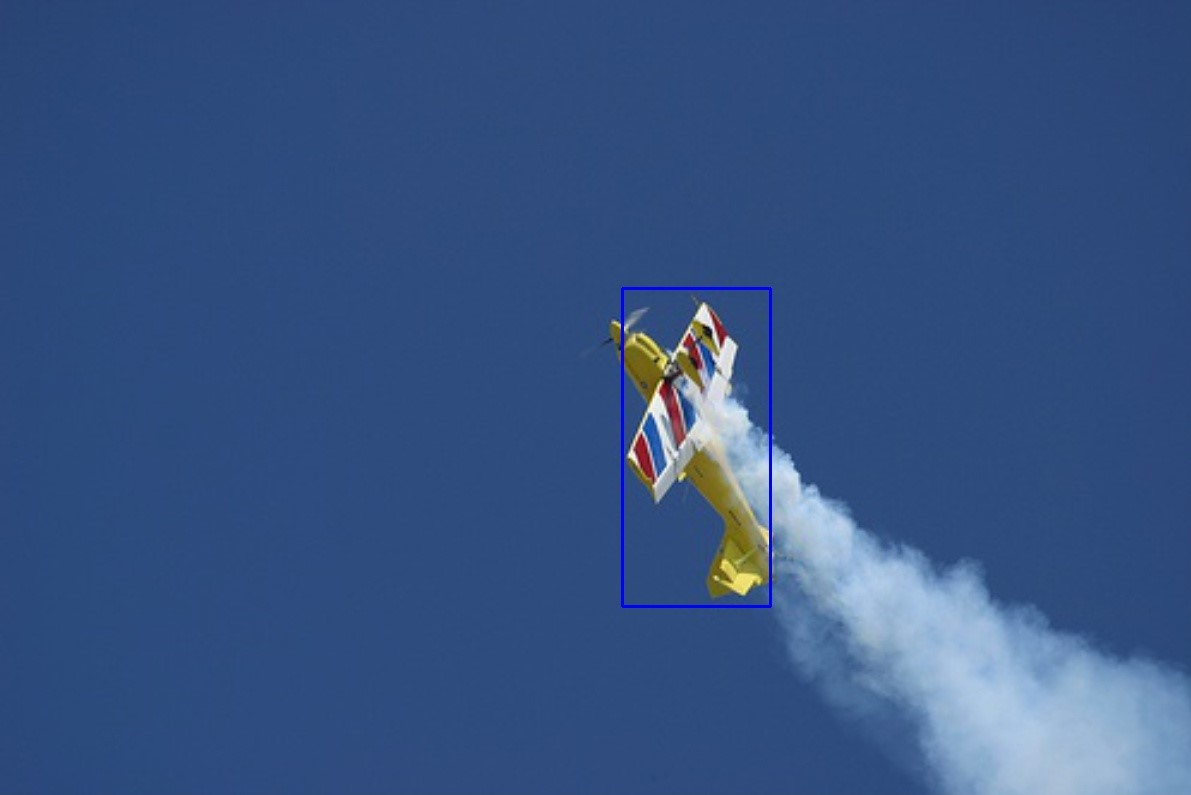}
\end{center}
\begin{center}
  \includegraphics[align=c,width=0.08\linewidth]{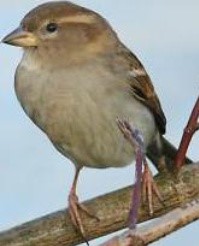}
  \hfill
  \includegraphics[align=c,width=0.15\linewidth]{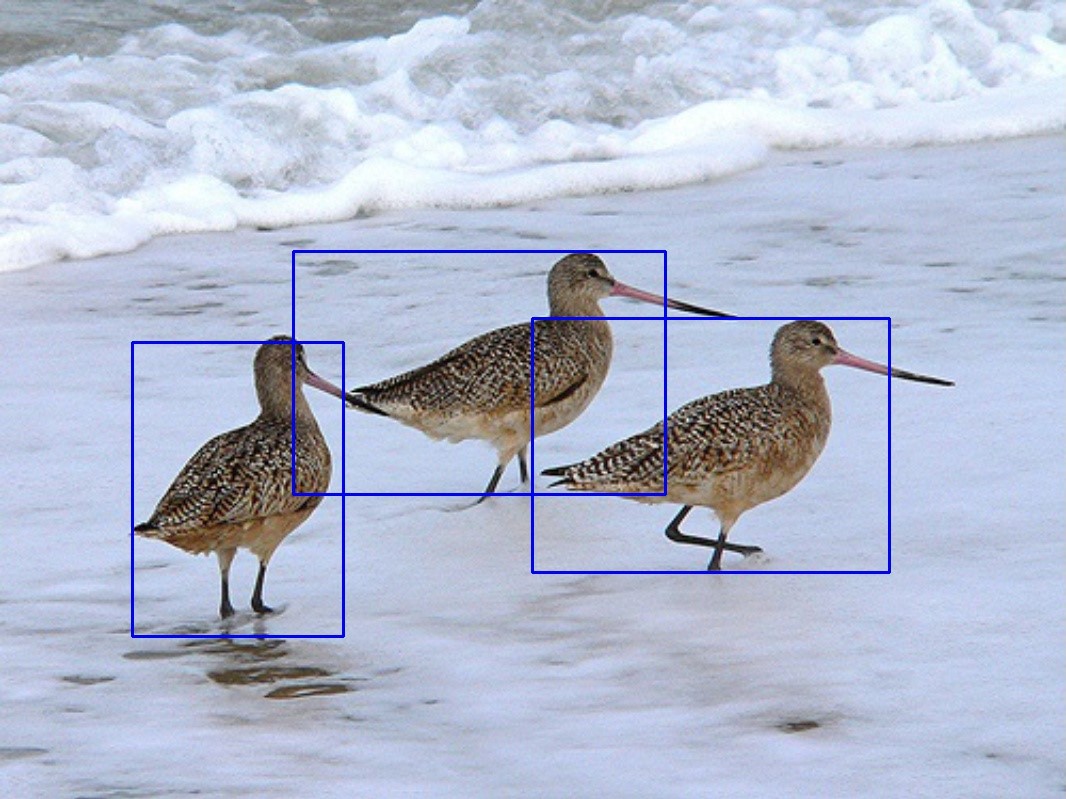}
  \includegraphics[align=c,width=0.15\linewidth]{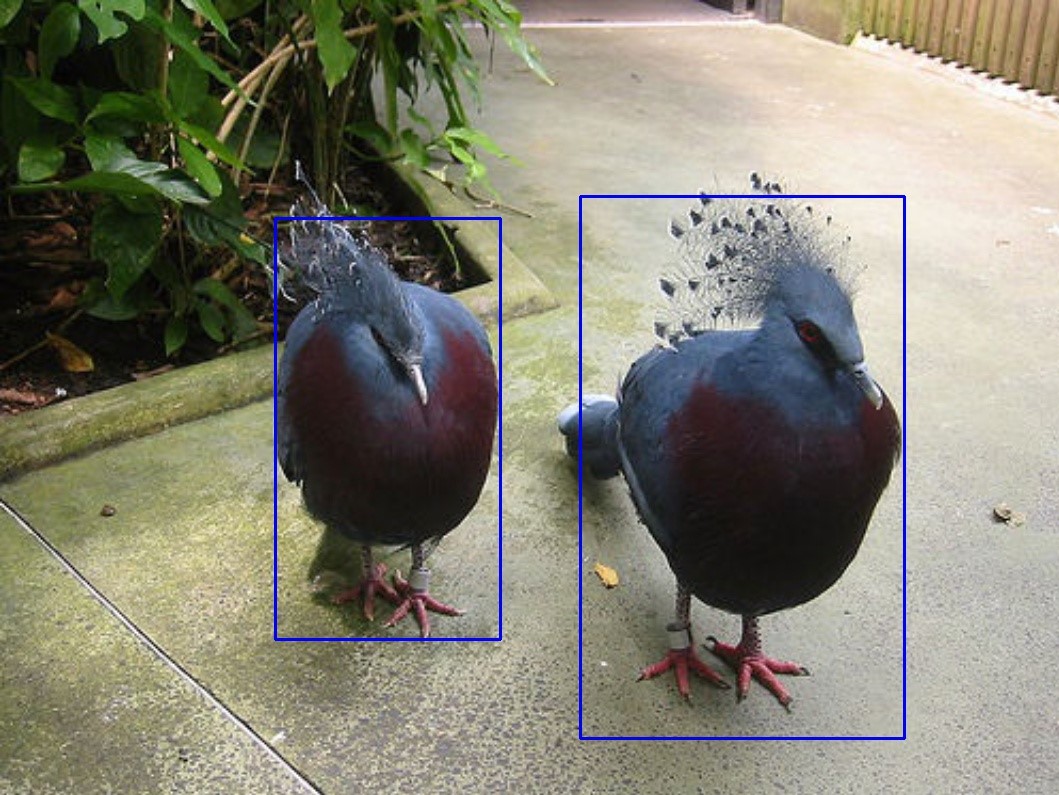}
  \includegraphics[align=c,width=0.15\linewidth]{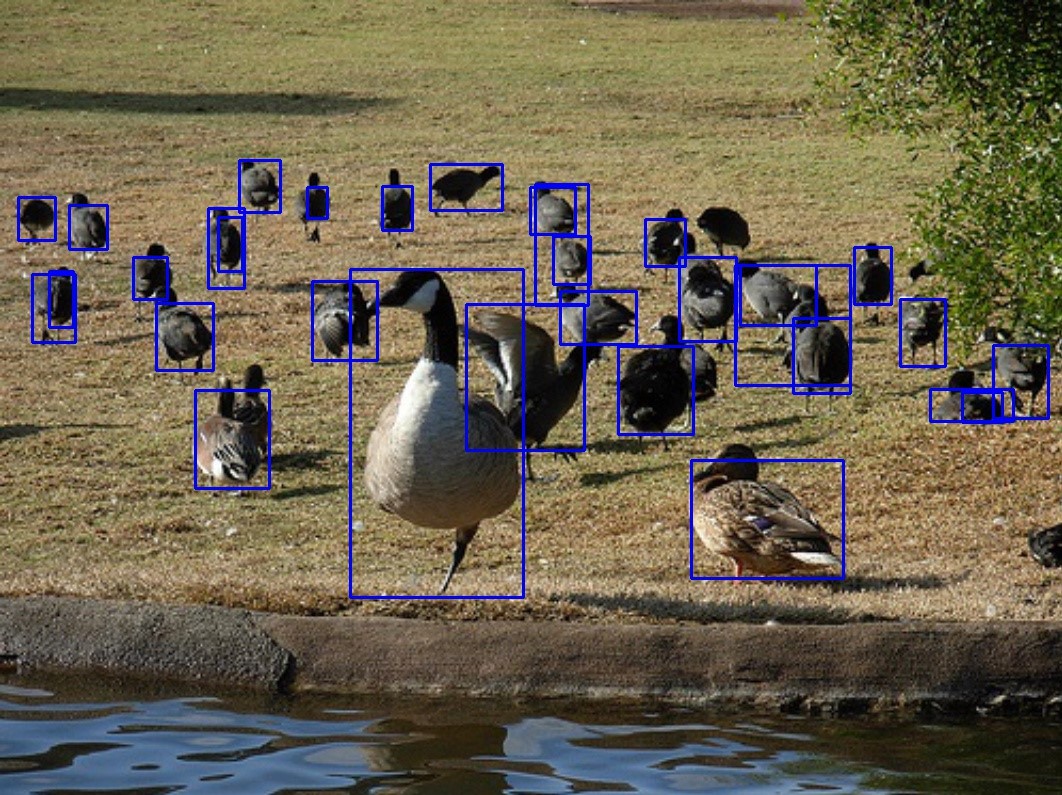}
  \includegraphics[align=c,width=0.15\linewidth]{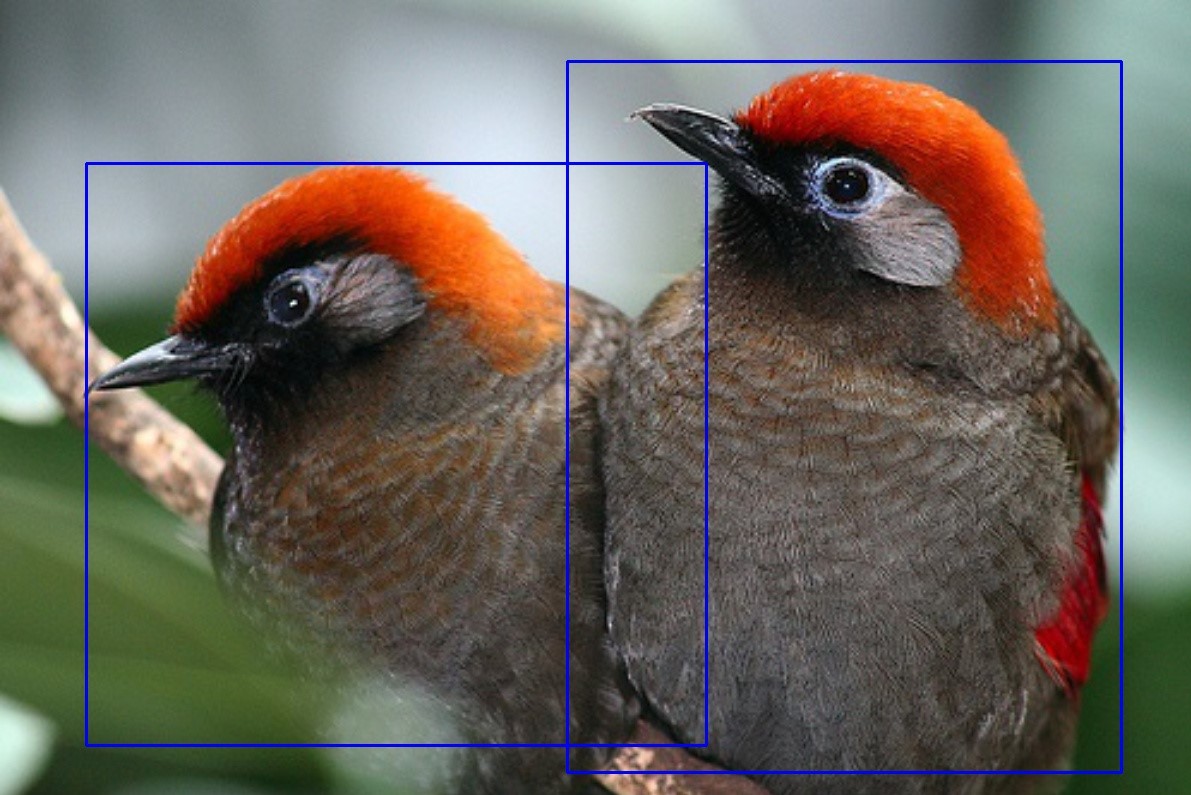}
  \includegraphics[align=c,width=0.15\linewidth]{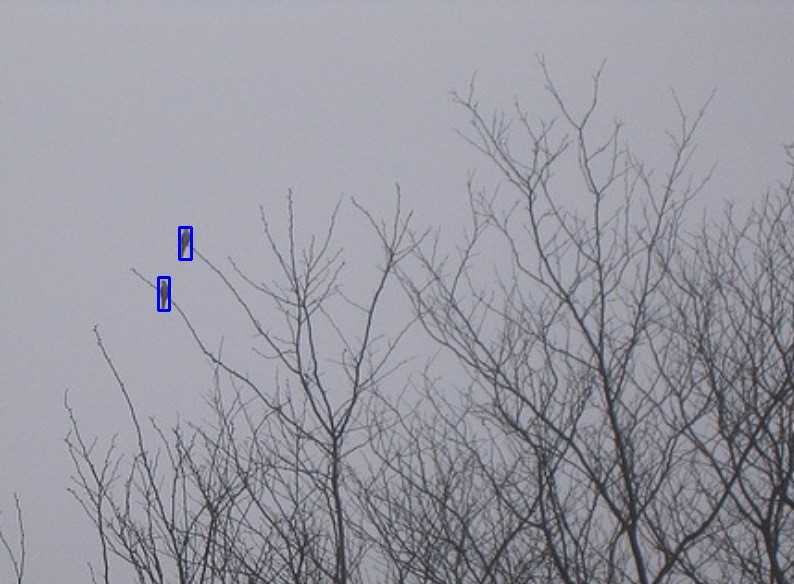}
\end{center}
\begin{center}
  \includegraphics[align=c,width=0.15\linewidth]{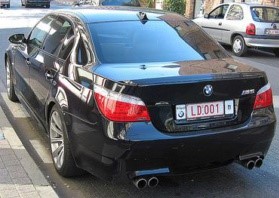}
  \hfill
  \includegraphics[align=c,width=0.15\linewidth]{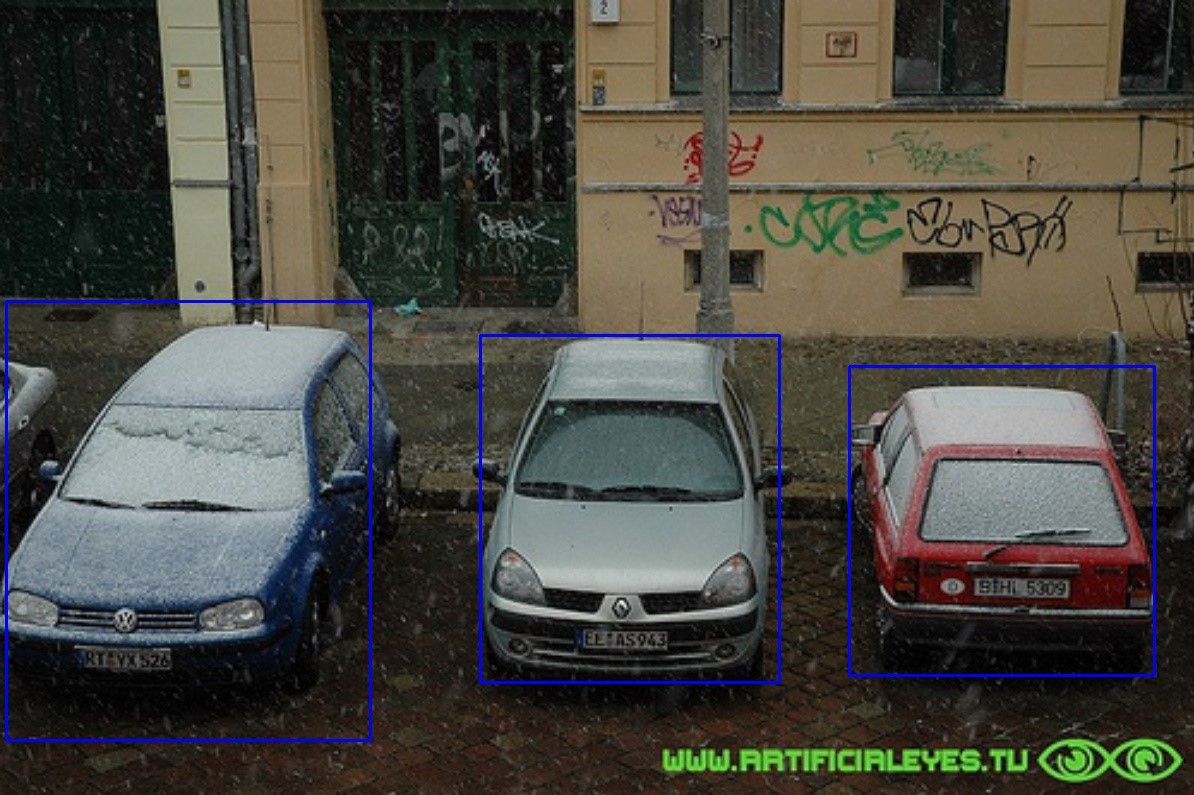}
  \includegraphics[align=c,width=0.15\linewidth]{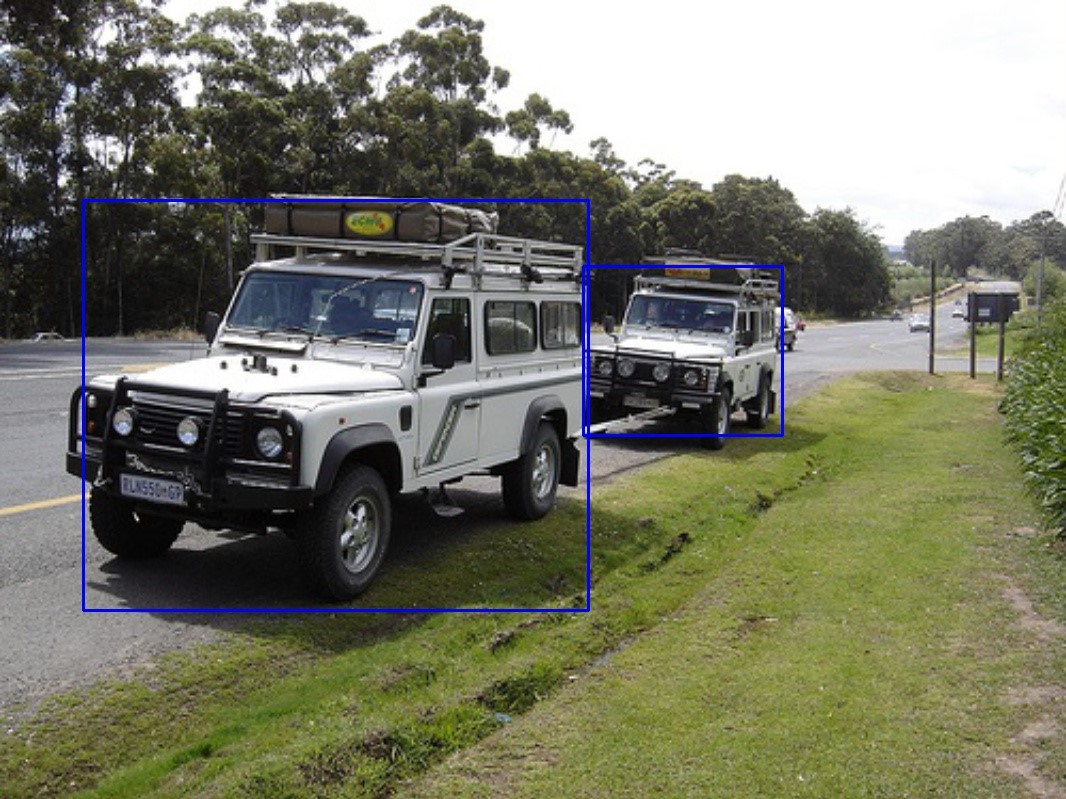}
  \includegraphics[align=c,width=0.15\linewidth]{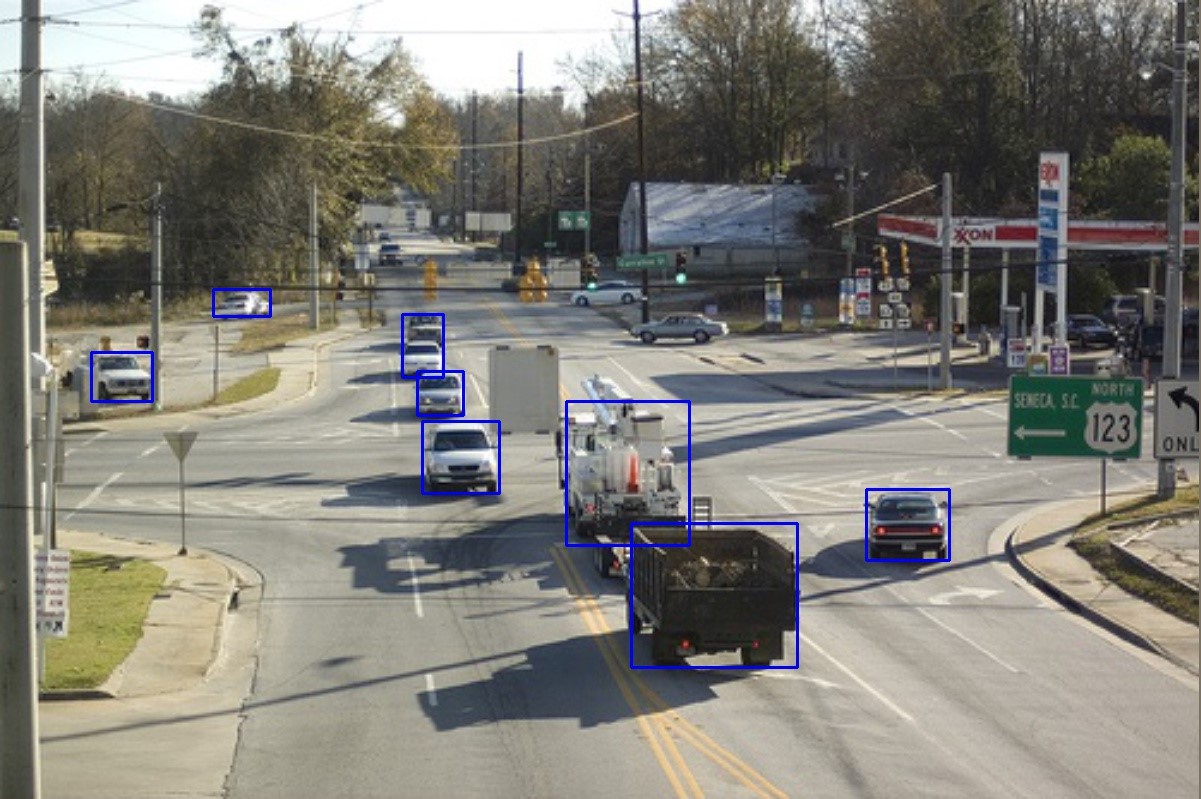}
  \includegraphics[align=c,width=0.15\linewidth]{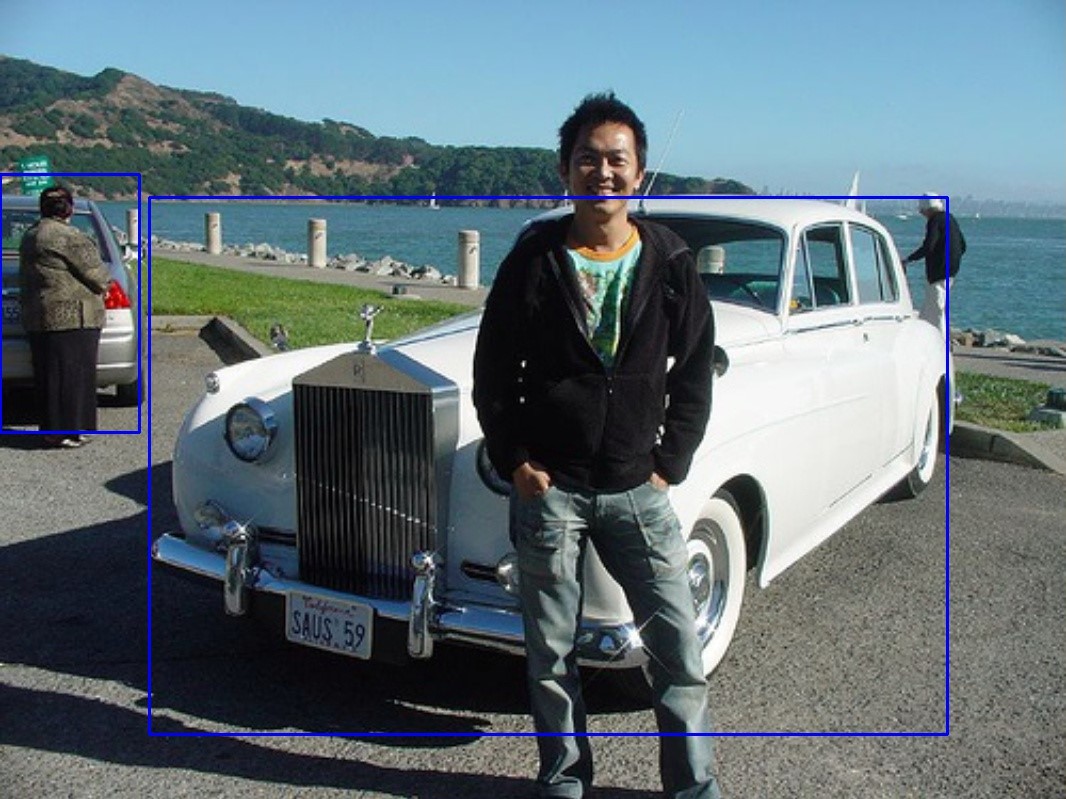}
  \includegraphics[align=c,width=0.15\linewidth]{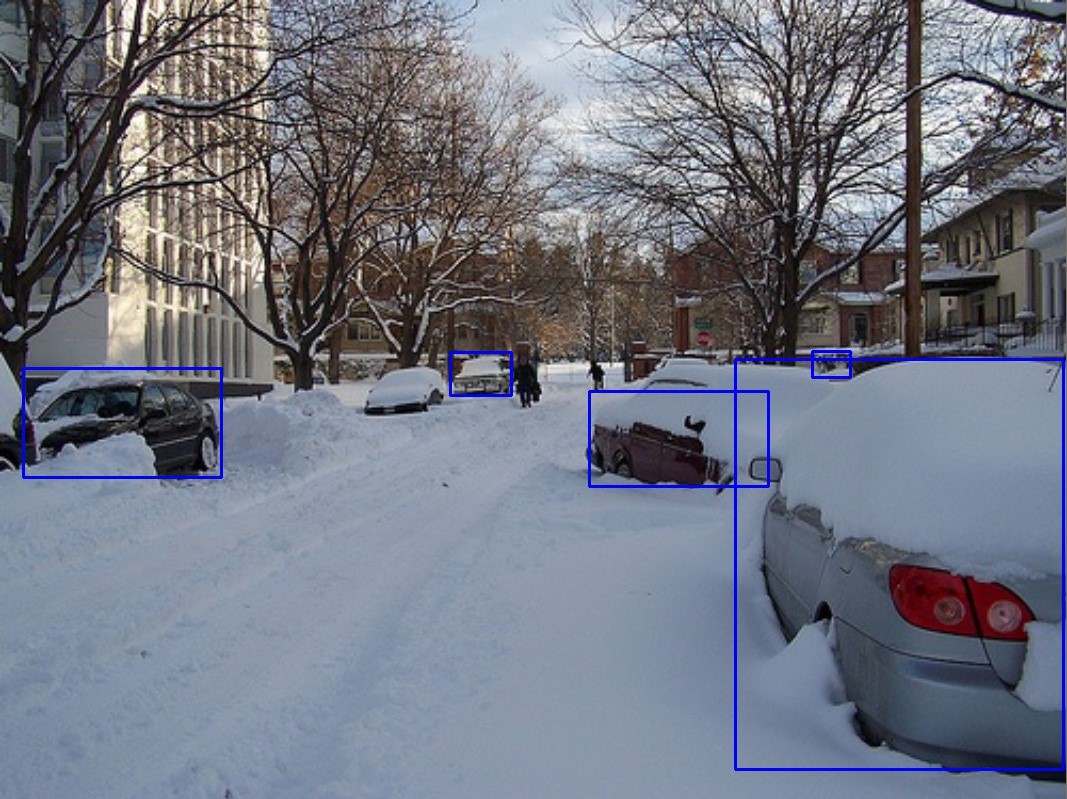}
\end{center}
\begin{center}
  \includegraphics[align=c,width=0.1\linewidth]{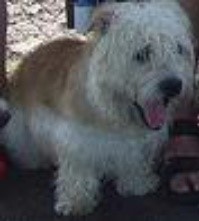}
  \hfill
  \includegraphics[align=c,width=0.15\linewidth]{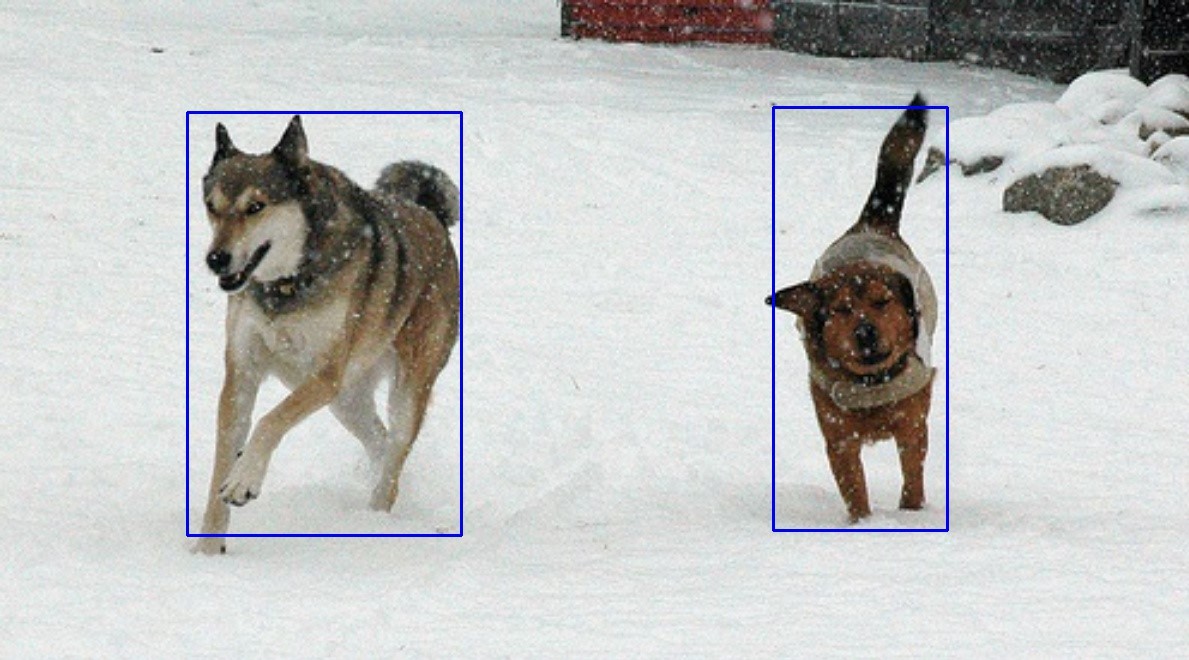}
  \includegraphics[align=c,width=0.15\linewidth]{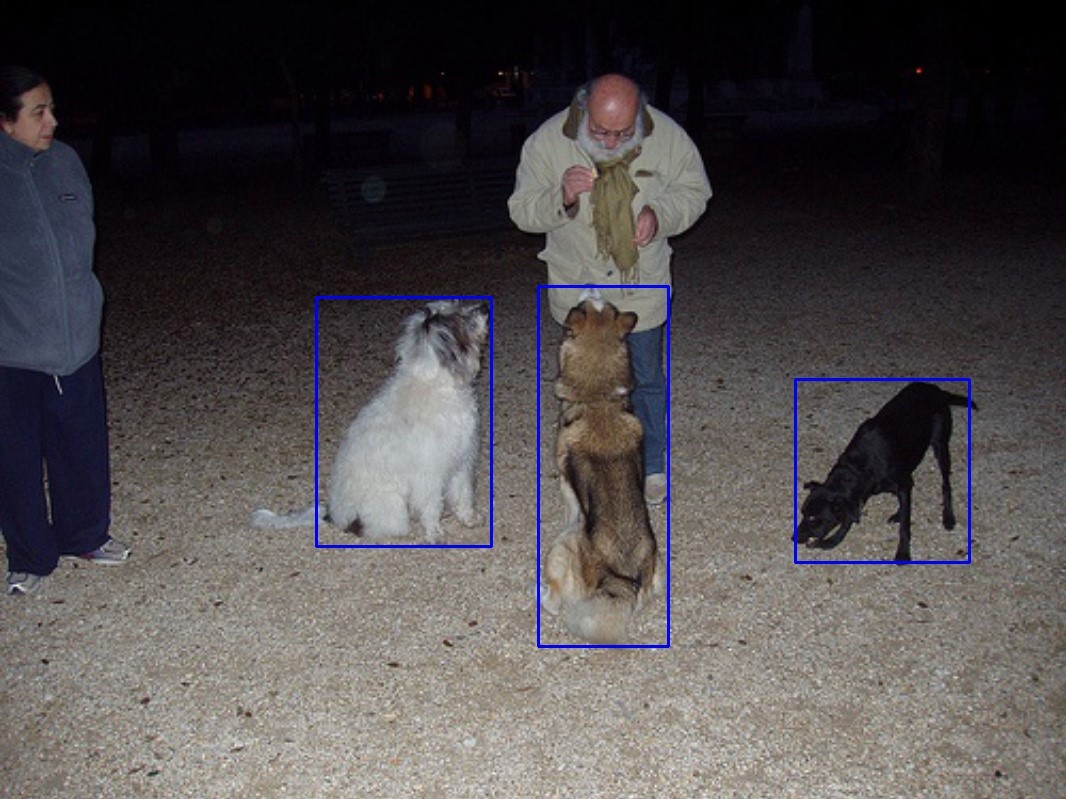}
  \includegraphics[align=c,width=0.15\linewidth]{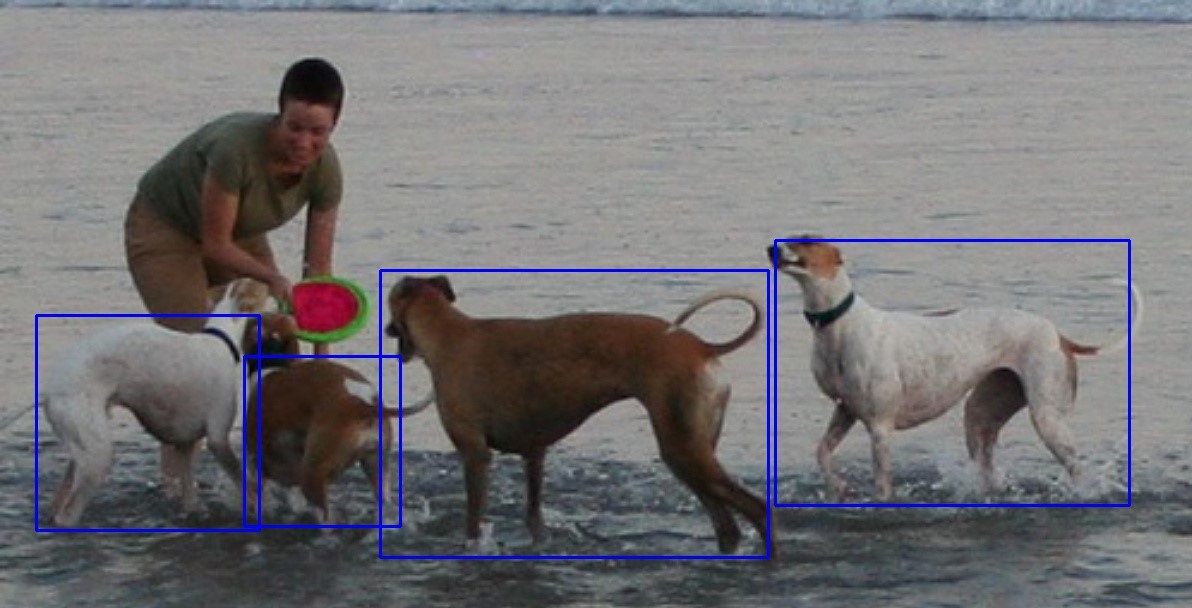}
  \includegraphics[align=c,width=0.15\linewidth]{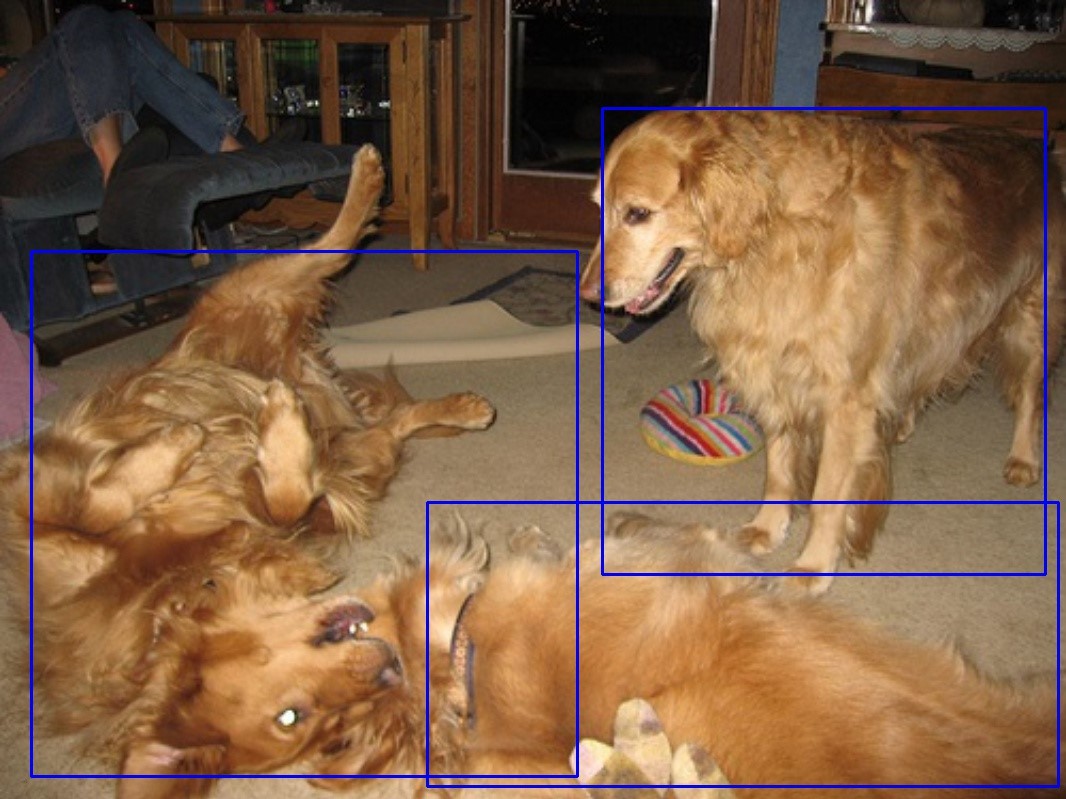}
  \includegraphics[align=c,width=0.15\linewidth]{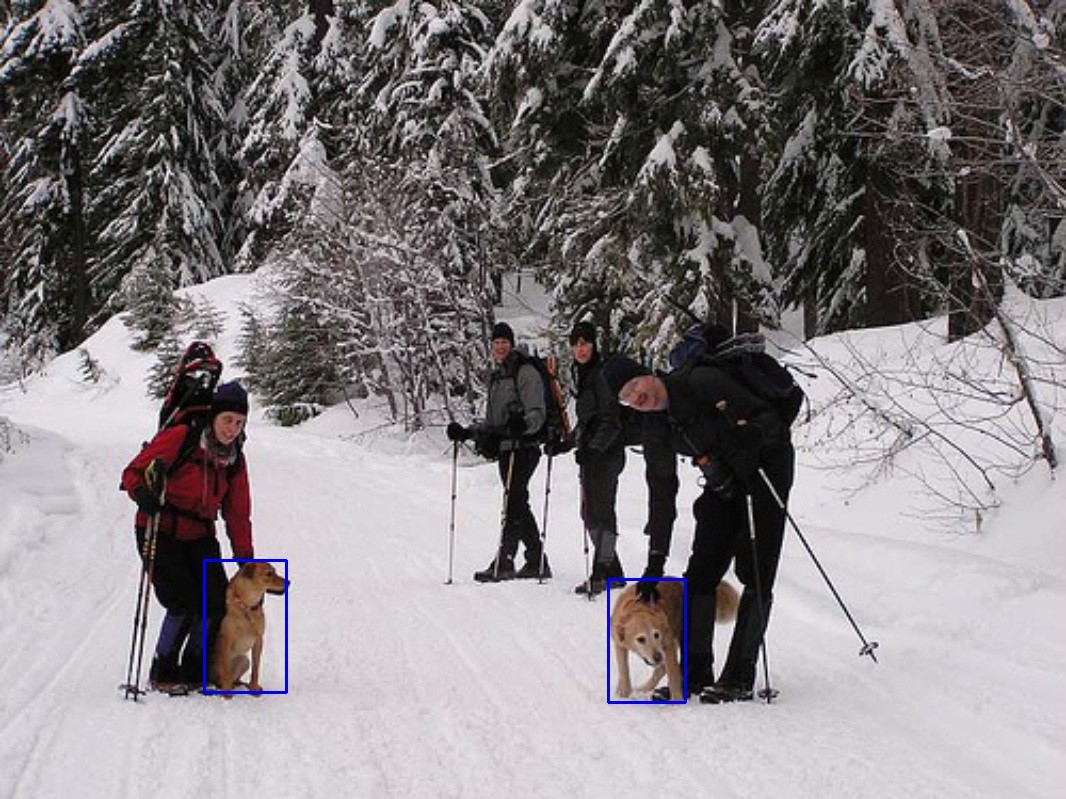}
\end{center}
\begin{center}
  \includegraphics[align=c,width=0.15\linewidth]{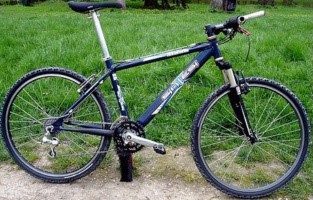}
  \hfill
  \includegraphics[align=c,width=0.15\linewidth]{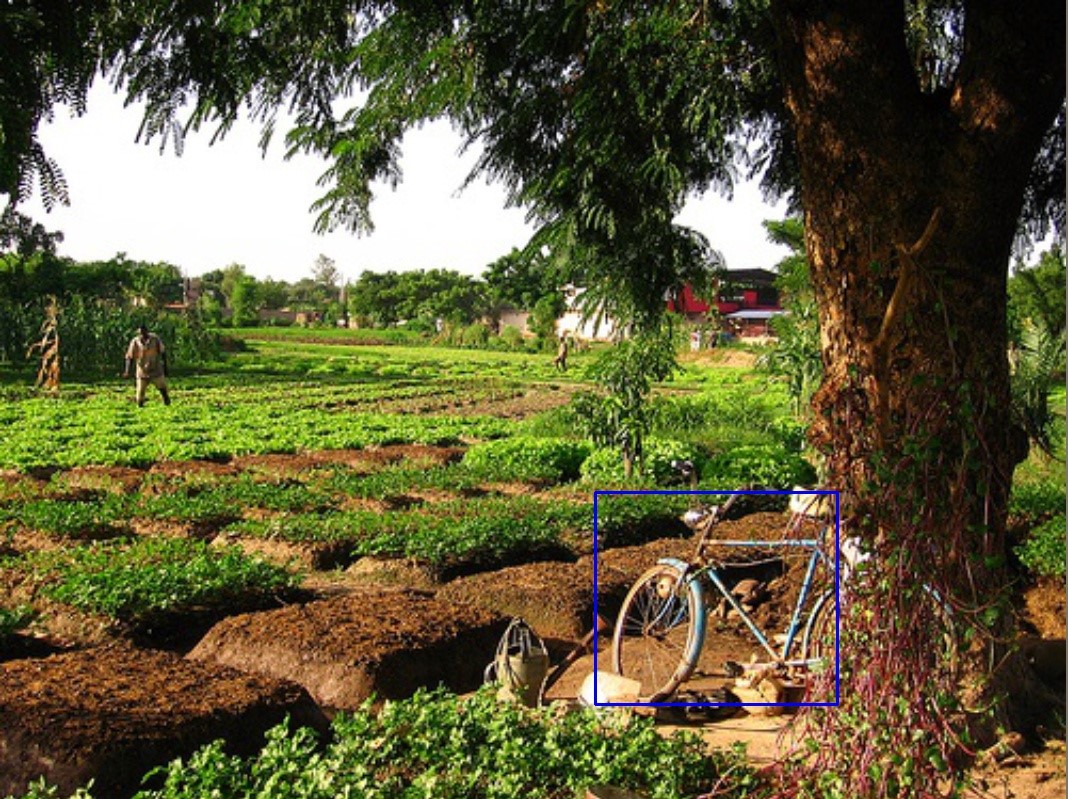}
  \includegraphics[align=c,width=0.15\linewidth]{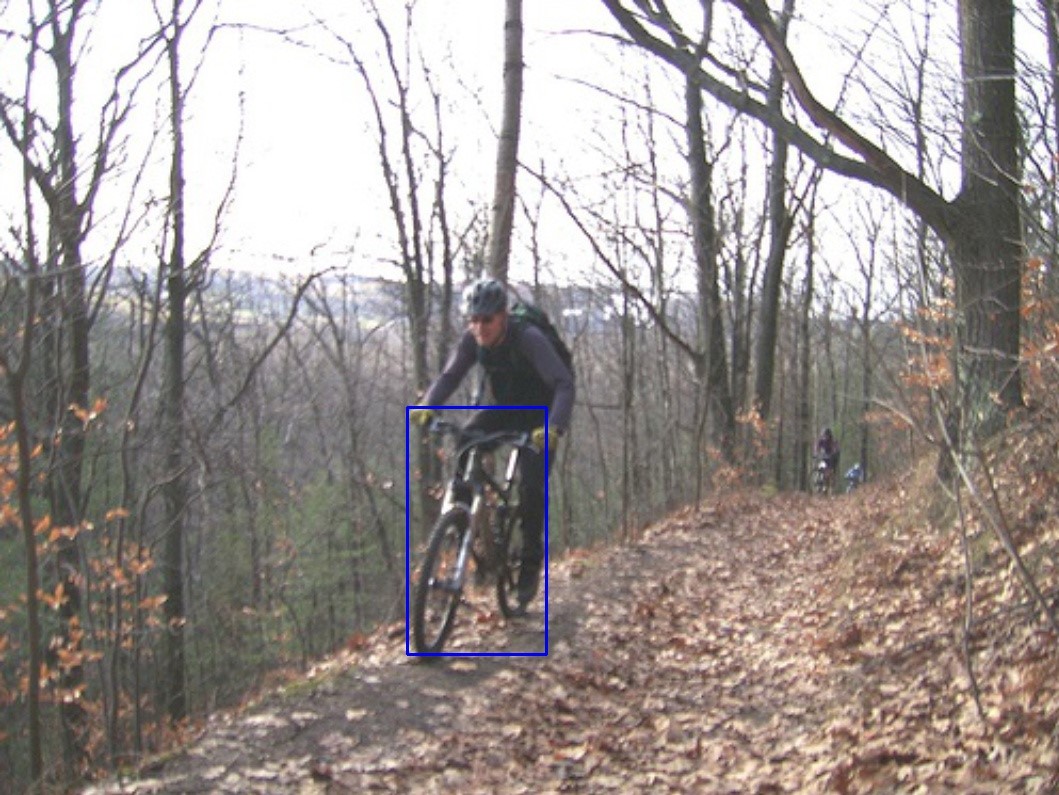}
  \includegraphics[align=c,width=0.15\linewidth]{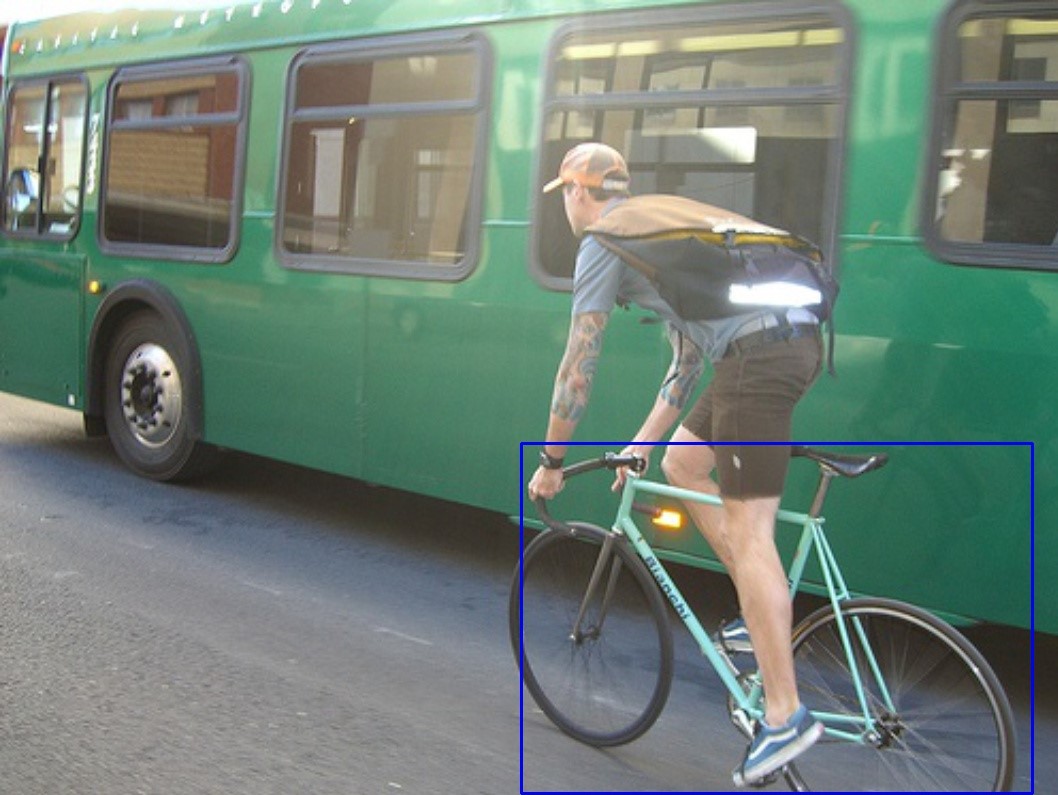}
  \includegraphics[align=c,width=0.15\linewidth]{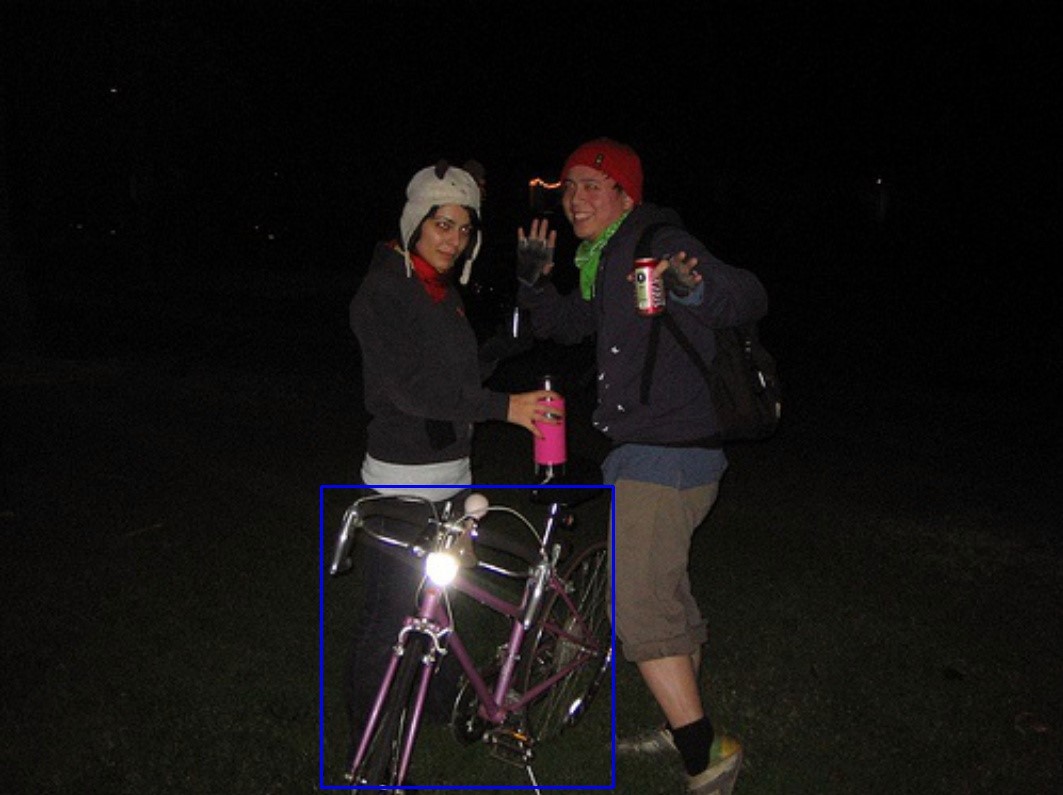}
  \includegraphics[align=c,width=0.11\linewidth]{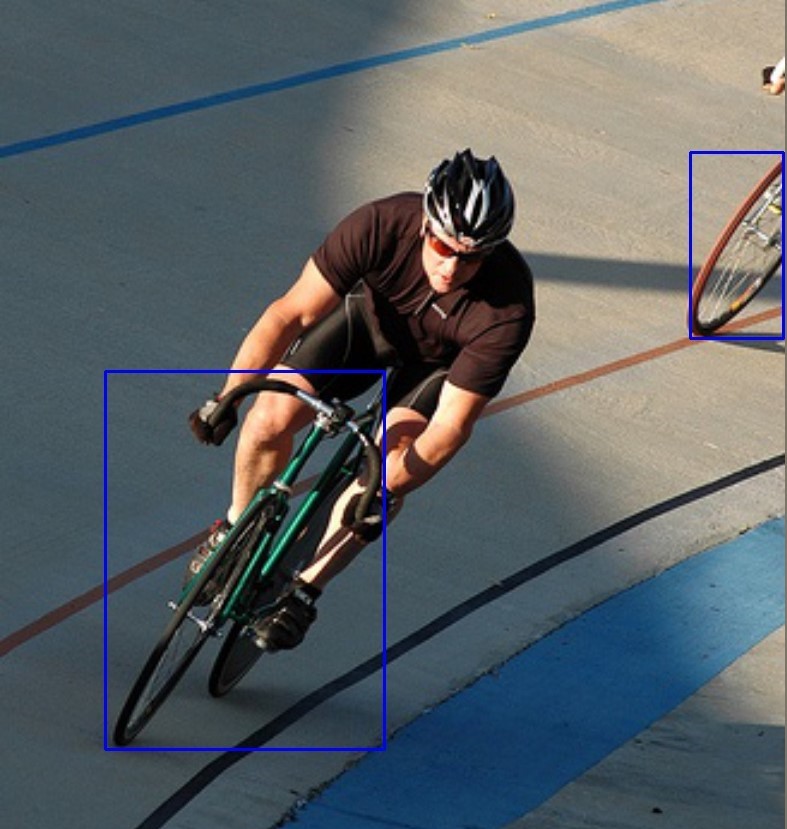}
\end{center}
\begin{center}
  \includegraphics[align=c,width=0.15\linewidth]{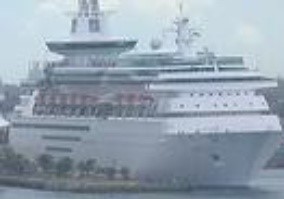}
  \hfill
  \includegraphics[align=c,width=0.15\linewidth]{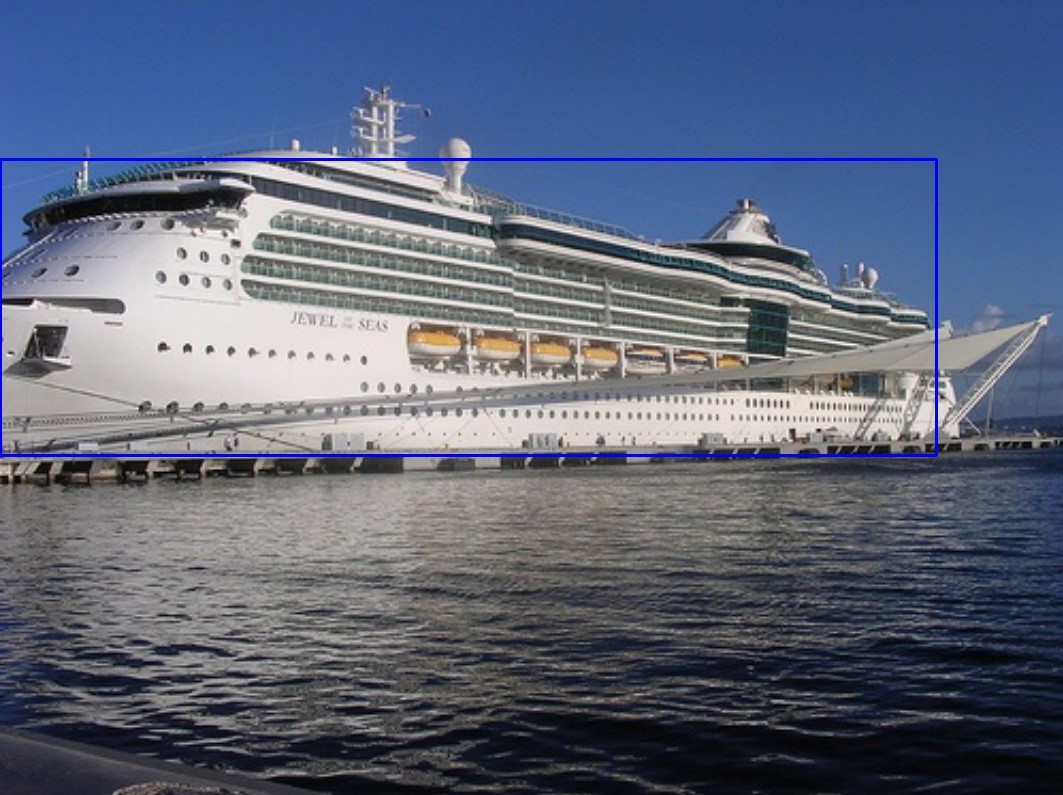}
  \includegraphics[align=c,width=0.15\linewidth]{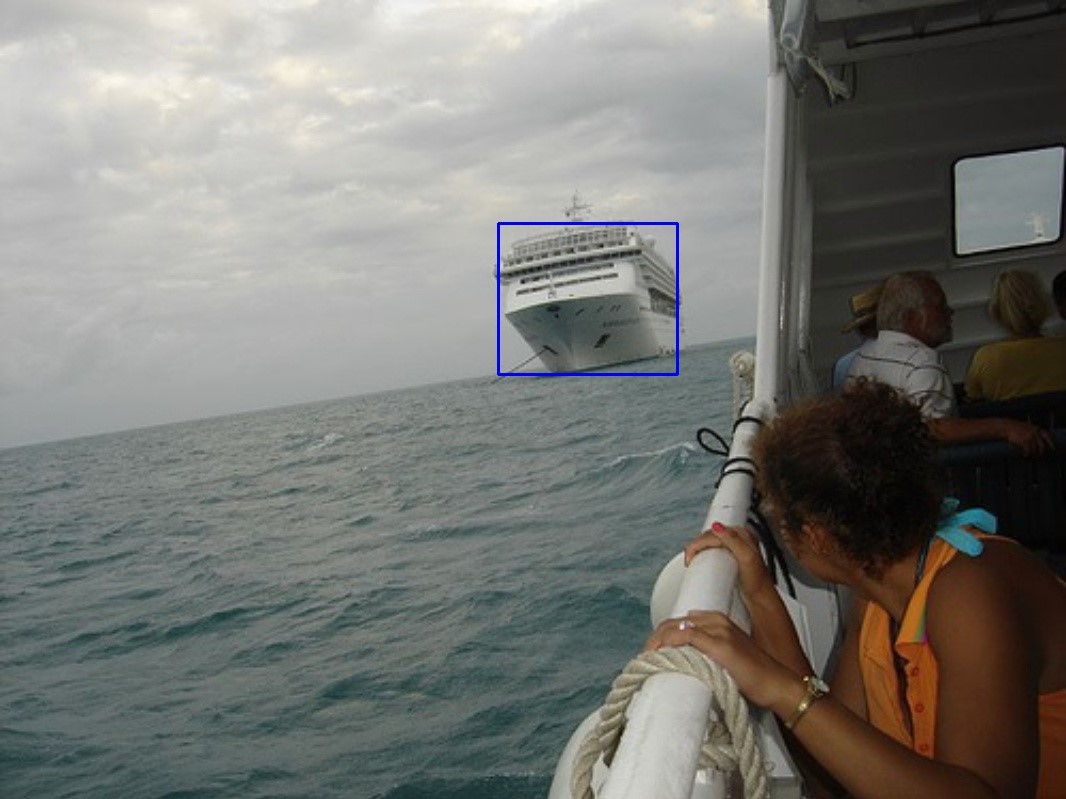}
  \includegraphics[align=c,width=0.15\linewidth]{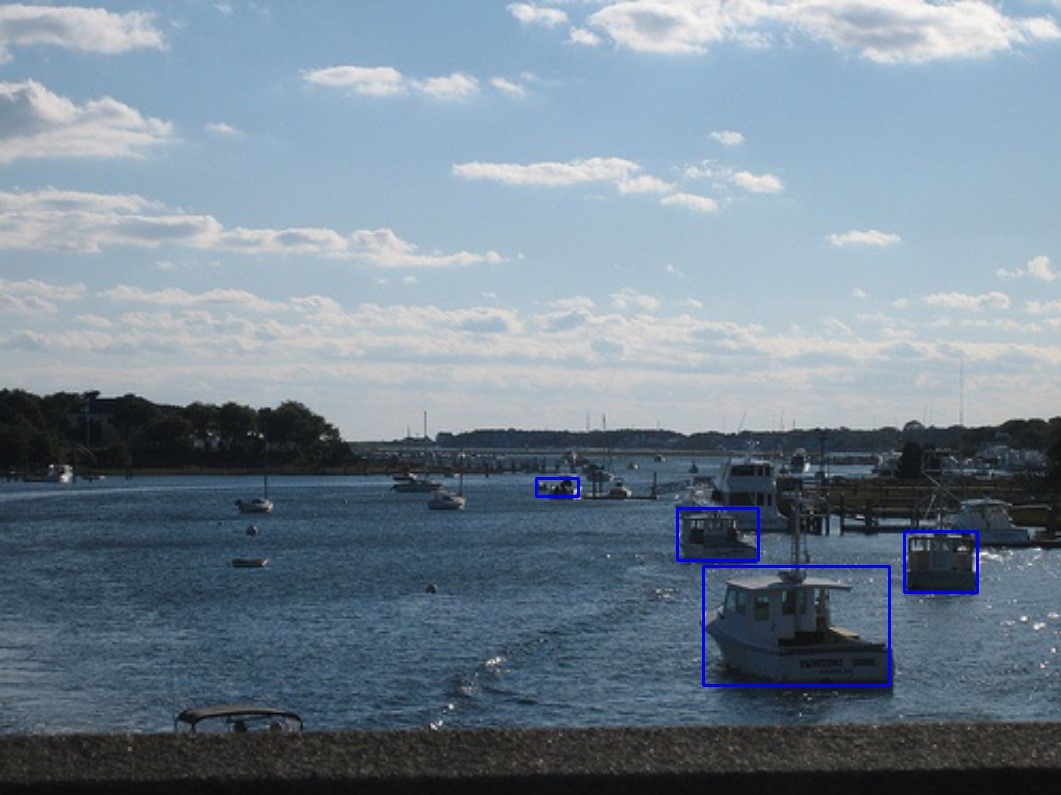}
  \includegraphics[align=c,width=0.15\linewidth]{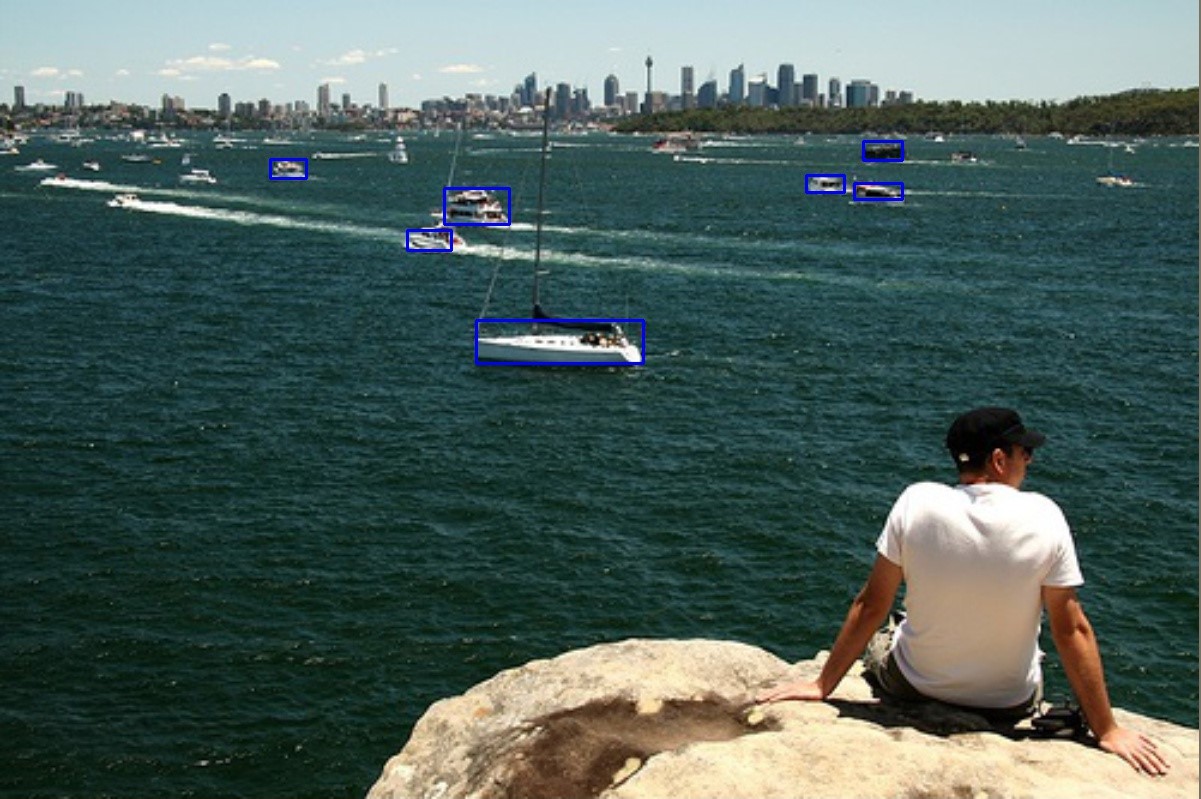}
  \includegraphics[align=c,width=0.15\linewidth]{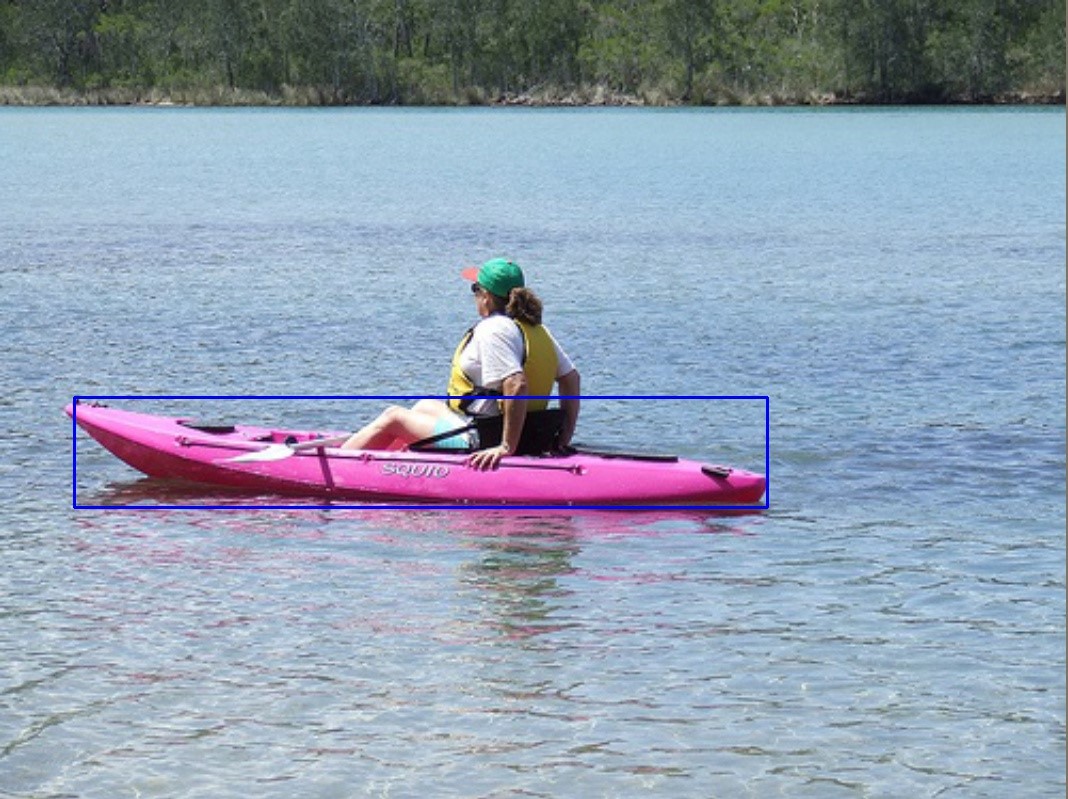}
\end{center}
\begin{center}
  \includegraphics[align=c,width=0.08\linewidth]{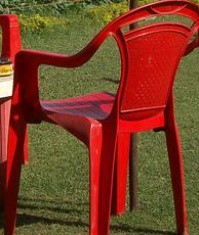}
  \hfill
  \includegraphics[align=c,width=0.15\linewidth]{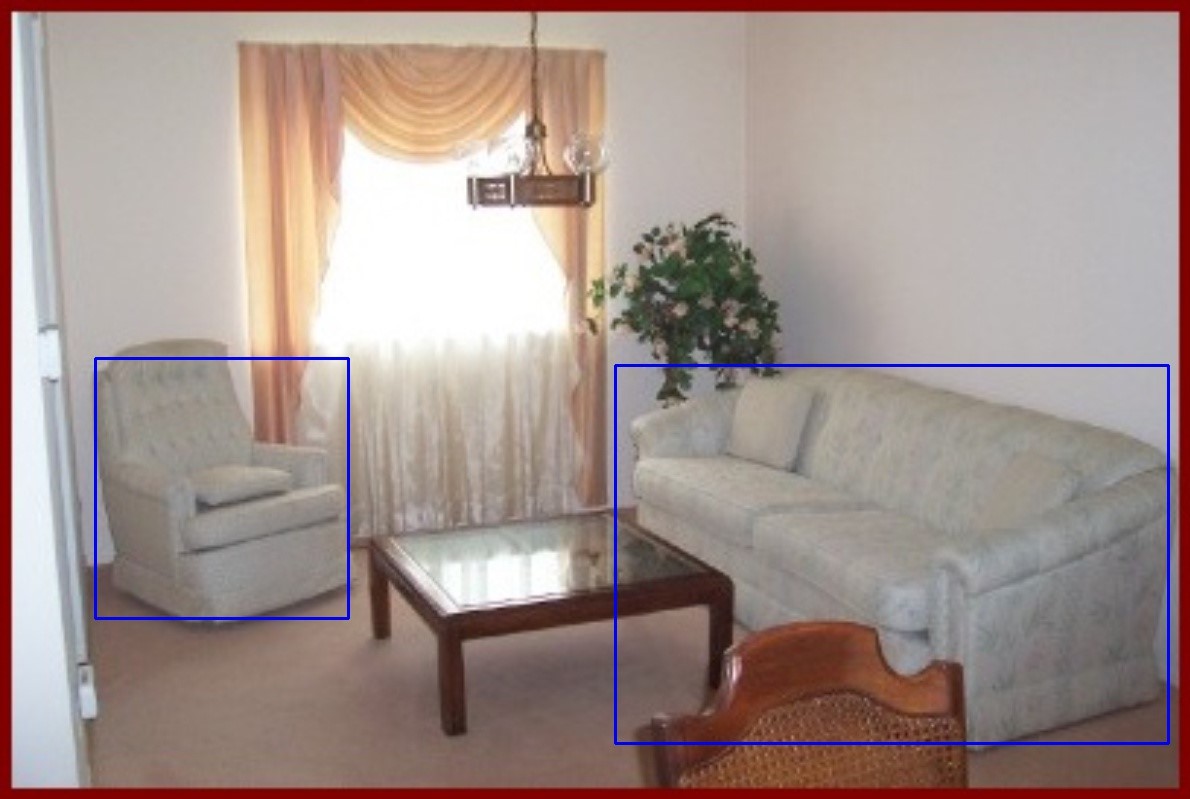}
  \includegraphics[align=c,width=0.15\linewidth]{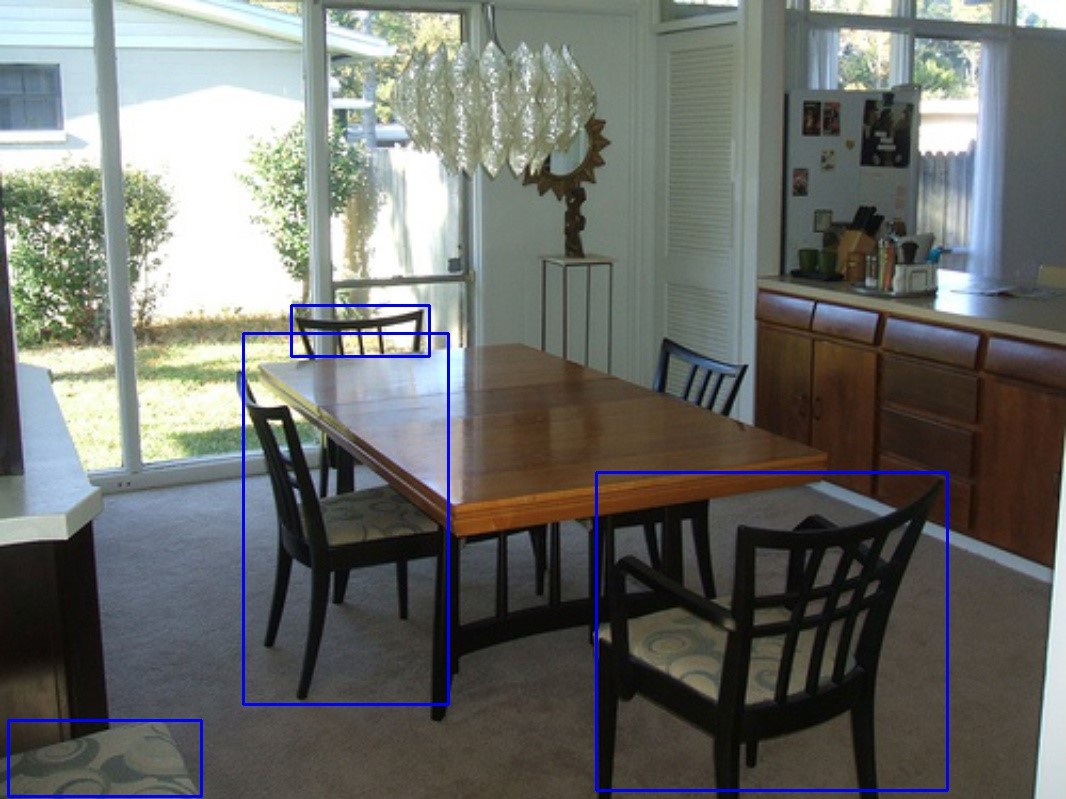}
  \includegraphics[align=c,width=0.15\linewidth]{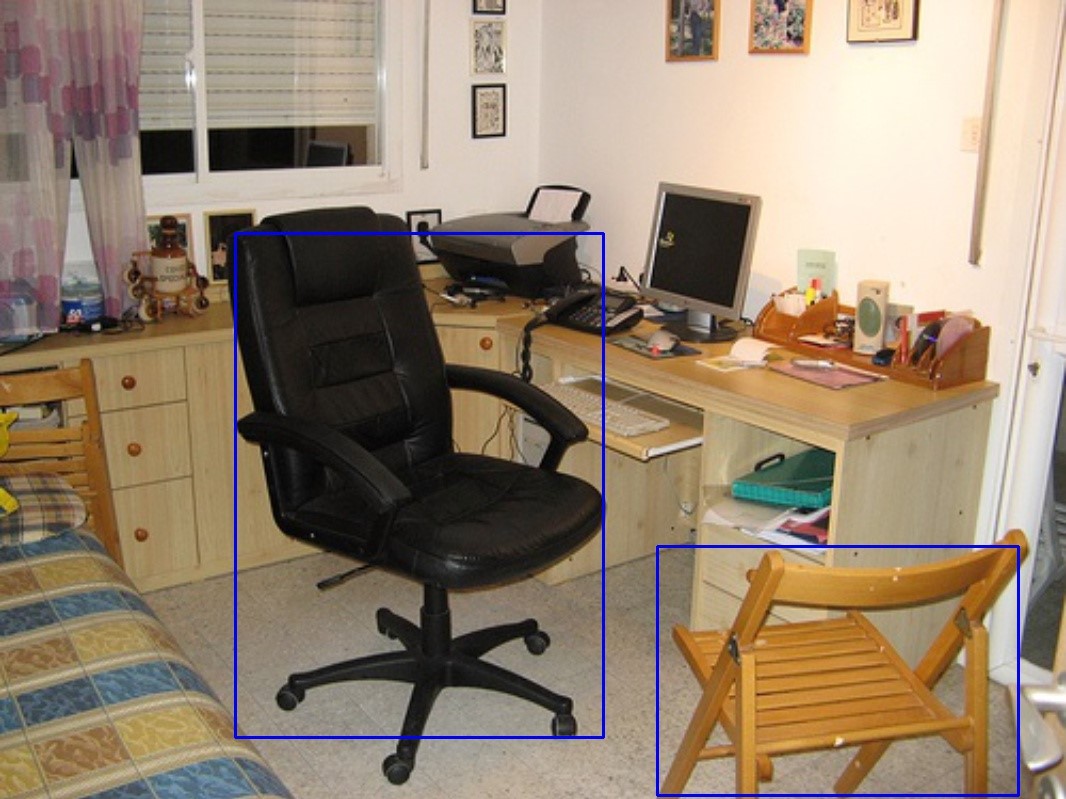}
  \includegraphics[align=c,width=0.15\linewidth]{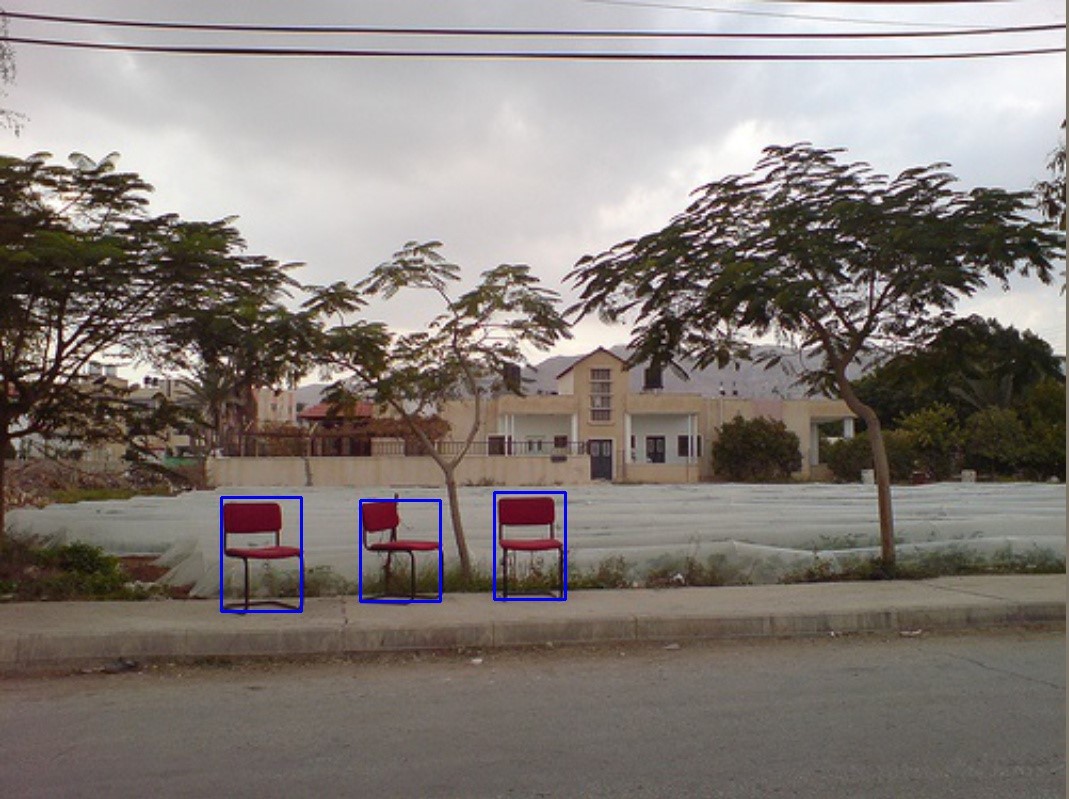}
  \includegraphics[align=c,width=0.15\linewidth]{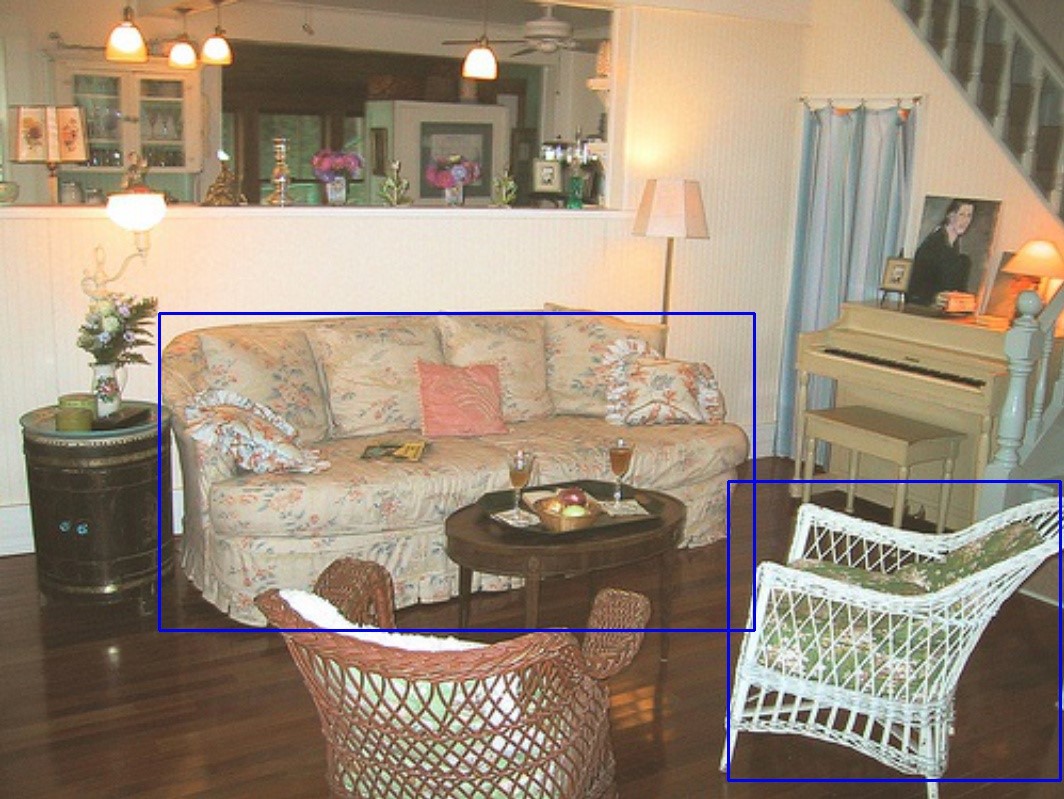}
\end{center}
\vspace{-0.1in}
  \caption{Qualitative results of our method on the PASCAL VOC 2007 test set. The leftmost column of images are support instance patches and the 5 other images on the right are query images.}
\label{qual}
\vspace{-0.2in}
\end{figure}



In this section, we compare our model with the previous work on one-shot object detection. The experiment setting of task~\RomanNumeralCaps{1} and~\RomanNumeralCaps{2} follows LSTD~\cite{lstd}, while task~\RomanNumeralCaps{3} follows CoAE~\cite{co-ae}. The three tasks span the most widely used object detection datasets, namely, COCO~\cite{coco}, PASCAL VOC~\cite{voc} and ImageNet-LOC~\cite{imagenet}. 

Specifically, in task \RomanNumeralCaps{1} the model is trained on the standard COCO~\cite{coco} 80 classes and evaluated on 50 other chosen classes in ImageNet2015~\cite{imagenet}; in task~\RomanNumeralCaps{2} the model is trained on 60 classes of COCO, excluding 20 classes contained in VOC2007~\cite{voc} as test classes. We believe these two cross-dataset tasks not only give a more accurate representation of the model's capacity by providing a sufficiently large training dataset, but are also more similar to the realistic use for one-shot object detection model. In task~\RomanNumeralCaps{3} the model is trained on 16 classes in VOC 2007 train \& val sets and VOC 2012 train \& val sets, and tested on the rest 4 classes in the VOC 2007 test set. Other settings exactly follows CoAE~\cite{co-ae}.  For a fair comparison, the Second Stage Knowledge Transfer is not used for our method in all comparisons. The training and test classes for each task are non-overlapping. \red{The evaluation metric is mainly mAP at 0.5 IoU following the convention of previous work, with mAP at 0.75 IoU in the ablation study to examine the detection quality more precisely.}
The detailed experiment setting and hyperparameters are available in the supplementary material.

Table~\ref{compWithLSTD} compares the performance of our method and LSTD and Repmet.
Under the 1-shot setting, our method outperforms LSTD by a large margin of 11.8\% and 12.1\% in AP\textsubscript{50}, also outperforming Repmet by 6.9\% in task~\RomanNumeralCaps{1}, showing that our method extracts the class-relevant information much more effectively from only 1 support instance patch, which is attributed to both the metric learning paradigm, our effective network architecture and our novel training strategies. 
Table~\ref{compWithCOAE} adds our method to Table~I of CoAE~\cite{co-ae}, which shows the comparison of our model against CoAE (and its baseline methods) on task~\RomanNumeralCaps{3}. Using the same Resnet-50 backbone, the same pretraining setting (1k) and the same training data, our method achieves a 0.9\% improvement on the mAP of unseen classes and 1\% improvement on the mAP of seen classes.
Table~\ref{perClass} shows the more detailed per-class performance of our method for task~II. 
\red{Qualitative results on task~II are available in Figure~\ref{qual}, 
which demonstrates that our model is robust to multiple instances, different perspectives, varying scales of objects, intra-class variance, and occlusion.  }








\subsection{Ablation study}\label{ablation}

\begin{table*}[t]
\begin{center}
\begin{adjustbox}{max width=0.99\linewidth}
\begin{tabular}{l|c|c|c||c|c|c}
\hline
Model&AP\textsubscript{50:95}&AP\textsubscript{50}&AP\textsubscript{75}&AR\textsubscript{1}&AR\textsubscript{10}&AR\textsubscript{100}\\
\hline
LSTD             & - & 34.0 & - & - & - & -  \\
Class-agnostic FCOS & 11.2 & 19.8\textsuperscript{\textbf{--14.2}} & 10.9 & 12.2 & 29.8 & 45.7  \\
Matching-FCOS    & 22.6\textsuperscript{\textbf{+11.4}} & 38.9\textsuperscript{\textbf{+19.1}} & 23.2\textsuperscript{\textbf{+12.3}} & 21.7\textsuperscript{\textbf{ +9.5}} & 44.7\textsuperscript{\textbf{+14.9}} & 58.9\textsuperscript{\textbf{+13.2}}  \\
Ours             & 24.8\textsuperscript{\textbf{ +2.2}} & 46.1\textsuperscript{\textbf{ +7.2}} & 23.7\textsuperscript{\textbf{ +0.5}} & 23.0\textsuperscript{\textbf{ +1.3}} & 42.5\textsuperscript{\textbf{ --2.2}} & 51.4\textsuperscript{\textbf{ --7.5}}  \\
\hline
\end{tabular}
\end{adjustbox}
\end{center}
\vspace{-0.05in}
\caption{  Ablation study of our network architecture with comparison to LSTD on task~\RomanNumeralCaps{2}. Superscripts shows the relative change w.r.t. the previous row.}
\vspace{-0.30in}
\label{structureAb}
\end{table*}

\noindent {\bf Matching-FCOS.~~}
Table~\ref{compRecall} shows that by introducing support information in the generation of object proposals, Matching-FCOS obtains a higher recall than the original RPN and FCOS. A high recall at 4000 proposals provides a high upper-bound for the second stage network, almost guaranteeing that the following classifier and box regressor will not be constrained by the false-negative proposals. Therefore, the second stage SARM can safely trade-off recall for improved precision, which is verified by the increase in AP and decrease in AR\textsubscript{10}, AR\textsubscript{100}. Matching-FCOS and our complete model are compared in Table~\ref{structureAb}. 

The effective integration of support information enables the Matching-FCOS to achieve much better results than the original FCOS trained class-agnostically. The Matching-FCOS alone already exceeds the state-of-the-art performance in the 1-shot setting by 4.9\%,  suggesting that it can be directly used as a one-stage one-shot object detector with the benefit of speed and simplicity.


\noindent {\bf Structure-Aware Relation Module.~~} The significance of a second stage classification and box regression step is shown in Table~\ref{structureAb}, where it contributes 7.2\% of the AP\textsubscript{50}.  Table~\ref{compHead} further justifies the effectiveness of the Structure-Aware Relation Module (SARM). Compared to a Global-average head, which globally averages both the support and query proposal features and processes them with fully-connected layers, and a trivial few-shot extension to a standard Faster R-CNN head, SARM achieves a better performance in both AP\textsubscript{50} and AP\textsubscript{75}. These experiments demonstrate that the more fine-grained one-shot classification and box regression at the second stage requires more detailed information in the spatial structure of the support and query features, which we aim to accomplish in our model by introducing SARM.

\begin{table}[t] 
\begin{center}
\begin{adjustbox}{max width=0.99\linewidth}
\begin{tabular}{l|c|c}
\hline
Training Strategy & AP\textsubscript{50} & AP\textsubscript{75} \\
\hline
Original model & 37.6   & 19.0 \\
+ Feature Similarity Mining & 43.4 \textsuperscript{\textbf{+5.8}}  & 22.8 \textsuperscript{\textbf{+3.8}} \\
+ Support Augmentation  & 45.3\textsuperscript{\textbf{+1.9}} & 23.4 \textsuperscript{\textbf{+0.6}} \\
+ GT-Curated Proposal  & 46.1\textsuperscript{\textbf{+0.8}} & 23.7 \textsuperscript{\textbf{+0.3}} \\
\hline
\end{tabular}
\end{adjustbox}
\end{center}
\vspace{-0.05in}
\caption{  Ablation study of our training strategies on task~II. Superscripts shows the relative change w.r.t. the previous row.}
\vspace{-0.15in}
\label{traingAb}
\end{table}


\noindent {\bf Reducing Intra-class Variance in Training.~~}
Table~\ref{traingAb} shows that feature similarity training improves AP\textsubscript{50} by a large margin of 5.8\%, validating that intra-class variance can impede training. 
Our feature similarity mining strategy reduces the noise of the training data by reducing intra-class variance 
which also helps to generalize to novel test classes better. Surprisingly, the simple horizontal flipping augmentation also improves AP\textsubscript{50} by a moderate 1.9\%. The potential reason for the improvement, as shown by the qualitative results, is that flipping augmentation compensates for the lack of such invariance in CNN, which allows the network to recognize a left-facing plane better provided by a right-facing support. 
Therefore, reducing intra-class variance in the training data may be instrumental in training a one-shot object detection model.

\begin{table}[t] 
\begin{center}
\begin{adjustbox}{max width=0.99\linewidth}
\begin{tabular}{c|l|c|c}
\hline
Index & Training Data & AP\textsubscript{50} & AP\textsubscript{75} \\
\hline
\RomanNumeralCaps{1} & COCO(60) & 46.1 & \textbf{23.7}  \\
\RomanNumeralCaps{2} &COCO(60) + FSS & 27.8  & 6.9  \\
\RomanNumeralCaps{3} &COCO(60) + FSS (SSKT)  & \textbf{48.9}  & 23.3  \\
\hline
\end{tabular}
\end{adjustbox}
\end{center}
\vspace{-0.05in}
\caption{  Comparison of detection performance with and without the Second Stage Knowledge Transfer (SSKT). Model~\RomanNumeralCaps{1} is trained on the COCO 60 classes according to task~II. Model~\RomanNumeralCaps{2} is trained directly on the COCO 60 classes and FSS~\cite{FSS1000}. Model~\RomanNumeralCaps{3} applies the Second Stage Knowledge Transfer with FSS apart from training on the COCO 60 classes. All models are tested on VOC2007 test set as in task~\RomanNumeralCaps{2}.}
\vspace{-0.35in}
\label{sskt}
\end{table}

\noindent {\bf Training SARM.~~}
Apart from the previous strategies which improve training of the entire model, we specifically propose two methods beneficial to the training of SARM at the second stage. As shown in Table~\ref{traingAb}, Ground Truth curated proposals alleviates the problem of low-quality proposals, improving the training of the SARM and improving the AP\textsubscript{50} by 0.8\%. Meanwhile, the Second Stage Knowledge Transfer can utilize effectively additional classification/segmentation datasets. As tabulated in Table~\ref{sskt}, directly training our model with the additional FSS has a detrimental effect on the performance, since the distribution of bounding boxes in the FSS dataset~\cite{FSS1000} is very different from that in the VOC test set. With the Second Stage Knowledge Transfer, the AP\textsubscript{50} of model~III achieves an improvement of 2.8\% 
and its AP\textsubscript{75} remains comparable to model~I. We believe the Second Stage Knowledge Transfer can effectively improve  classification capability of SARM by introducing more variety with little to no damage to its box regression performance, especially compared to model~II where the FSS data is added naively and both AP\textsubscript{50} and AP\textsubscript{75} decrease drastically.




\vspace{-0.2in}

\section{Conclusion and Future Work}
\vspace{-0.1in}

We propose a novel two-stage object detection model for one-shot object detection, which consists of Matching-FCOS in the first stage to obtain a high recall, and SARM in the second stage to better classify and localize the target from object proposals, eliminating the need to fine-tune the model on novel class support images when combined. We also propose the following novel training strategies. To reduce the noise introduced by intra-class variance in training support images, we propose Query-Support Feature Similarity Mining with support augmentation; Ground Truth Curated proposals and Second Stage Knowledge Transfer to better train the second stage SARM. Our model performs exceptionally on different cross-dataset one-shot object detection tasks, surpassing the state-of-the-art by a large margin. We perform an ablation study to illustrate the effect of different design decisions in architecture and training strategies. Finally, to reduce class ambiguity and further improve the detection performance, more support instance patches for each class can be introduced, so the effective combination of information from multiple support patches is worthwhile future work. Also, although not required for our method, further performance improvement can be obtained by strategically fine-tuning our model on support patches.

{\small
\bibliographystyle{splncs04}
\bibliography{egbib}
}

\end{document}